\documentclass[oneside,onecolumn]{article}

\usepackage{blindtext} % Package to generate dummy text throughout this template 
\usepackage[sc]{mathpazo} % Use the Palatino font
\usepackage[T1]{fontenc} % Use 8-bit encoding that has 256 glyphs
\usepackage{microtype} % Slightly tweak font spacing for aesthetics

\usepackage[english]{babel} % Language hyphenation and typographical rules

\usepackage[hmarginratio=1:1,top=32mm,columnsep=20pt]{geometry} % Document margins
\usepackage[hang, small,labelfont=bf,up,textfont=it,up]{caption} % Custom captions under/above floats in tables or figures
\usepackage{booktabs} % Horizontal rules in tables

\usepackage{lettrine} % The lettrine is the first enlarged letter at the beginning of the text

\usepackage{enumitem} % Customized lists
\setlist[itemize]{noitemsep} % Make itemize lists more compact

\usepackage{abstract} % Allows abstract customization
\usepackage{titlesec} % Allows customization of titles
\titleformat{\section}[block]{\large\scshape\centering}{\thesection.}{1em}{} % Change the look of the section titles
\titleformat{\subsection}[block]{\large}{\thesubsection.}{1em}{} % Change the look of the section titles

\usepackage{fancyhdr} % Headers and footers
\usepackage{titling} % Customizing the title section

\usepackage[hidelinks,colorlinks=true,linkcolor=blue,citecolor=blue]{hyperref} % For hyperlinks in the PDF
\usepackage[utf8]{inputenc}
\usepackage[T1]{fontenc}
\usepackage{natbib}
\usepackage{subcaption}
\usepackage[utf8]{inputenc}
\usepackage{amsmath, amsthm, amsfonts, amssymb} 
\usepackage{mathrsfs}  
\usepackage{blkarray}
\usepackage{caption}
\usepackage{subcaption}
\usepackage{multirow}
\usepackage[switch]{lineno}
\usepackage{xcolor}
\usepackage{comment}
\usepackage{arydshln}
\usepackage{mathtools}
\DeclareUnicodeCharacter{2212}{-}

\newcommand{\absdiv}[1]{%
  \par\addvspace{.5\baselineskip}% adjust to suit
  \noindent\textbf{#1}\quad\ignorespaces
}

\title{\LARGE  \textbf{EXPLAINABLE MACHINE LEARNING FOR PREDICTING HOMICIDE CLEARANCE \\IN THE UNITED STATES}} % Article title
\author{%
\textsc{Gian Maria Campedelli}\thanks{Corresponding author} \\[0.5ex] % Your name
\normalsize Department of Sociology and Social Research - University of Trento, Italy \\ % Your institution
\normalsize \href{mailto:gianmaria.campedelli@unitn.it}{gianmaria.campedelli@unitn.it} % Your email address
}
\date{February 2022\\
\color{purple} pre-print - this is the accepted version of the paper published in the Journal of Criminal Justice. \\ DOI: \href{https://doi.org/10.1016/j.jcrimjus.2022.101898}{https://doi.org/10.1016/j.jcrimjus.2022.101898}} % Leave empty to omit a date
\renewcommand{\maketitlehookd}{%
\begin{abstract}
\absdiv{Purpose}
To explore the potential of Explainable Machine Learning in the prediction and detection of drivers of cleared homicides at the national- and state-levels in the United States.
\absdiv{Methods}
First, nine algorithmic approaches are compared to assess the best performance in predicting cleared homicides country-wise, using data from the Murder Accountability Project. The most accurate algorithm among all (XGBoost) is then used for predicting clearance outcomes state-wise. Second, SHAP, a framework for Explainable Artificial Intelligence, is employed to capture the most important features in explaining clearance patterns both at the national and state levels. 
\absdiv{Results}
At the national level, XGBoost demonstrates to achieve the best performance overall. Substantial predictive variability is detected state-wise. In terms of explainability, SHAP highlights the relevance of several features in consistently predicting investigation outcomes. These include homicide circumstances, weapons, victims' sex and race, as well as number of involved offenders and victims.
\absdiv{Conclusions}
Explainable Machine Learning demonstrates to be a helpful framework for predicting homicide clearance. SHAP outcomes suggest a more organic integration of the two theoretical perspectives emerged in the literature. Furthermore, jurisdictional heterogeneity highlights the importance of developing ad hoc state-level strategies to improve police performance in clearing homicides.
\vspace{0.5cm}

\textbf{Keywords:} Murder Accountability Project; Explainable Artificial Intelligence; SHAP; Algorithmic Criminology; Homicide
\end{abstract}
}

\begin{document}

% Print the title
\maketitle

%----------------------------------------------------------------------------------------
%	ARTICLE CONTENTS
%----------------------------------------------------------------------------------------

\section*{Introduction}
During 2020 and 2021, the United States has experienced a large surge in homicides across the entire country \citep{AsherMurderRoseAlmost2021, Thebault2020wasdeadliest2021}, fueling intense public debates around the nature and consequences of the violent trend. While causes of this peak are yet to be determined, such a trend aggravated a national problem that places the United States as one of the Western countries with the highest rates of deaths caused by homicide. 

\begin{comment}
Data from the National Center for Health Statistics of the Center for Disease Control and Prevention, for instance, report that in 2019  states like Mississippi, Louisiana, and Alabama had homicide death rates that rank them around the 30th position in the global list of countries with highest rates of lethal violence worldwide \citep{WorldBankIntentionalhomicides1002018, NationalCenterforHealthStatisticsHomicideMortalityState2019}. 
\end{comment}
The direct human, social, economic, and political costs of homicidal violence \textit{per se} are, however, not the only ones impacting on US society. In fact, clearance rates for homicides in the United States have been declining over the years, implicating further issues to the problem of homicide prevalence \citep{RiedelDeclineArrestClearances1999,OuseyKnowUnknownDecline2010, CouncilonCriminalJusticeHomicideTrendsWhat2021}. 

Low levels of clearance rates for homicides represent the rubbed salt in the deep wound of homicidal violence plaguing the United States. Uncleared homicides substantially undermine the sense of safety of victims' families and communities in which victims were embedded, reinforcing trauma and stress, furthermore corroding trust between citizens and institutions, and especially law enforcement and criminal justice ones \citep{RegoecziUnclearedHomicidesCanada2000, BragaImprovingPoliceClearance2021}.

Two major conceptual perspectives dominate the criminological debate regarding the underlying explanations for homicide clearances. First, Black's theory of law \cite{BlackBehaviorLaw1976} and the "discretionary"  perspective \citep{PaternosterProsecutorialDiscretionRequesting1984, PetersonChangingConceptionsRace1984} posit that police uses discretion in clearing homicides based on the characteristics of the victims, like the victim's race or age. Second, the nondiscretionary perspective, championed by \cite{WolfgangPatternsCriminalHomicide1958}, \cite{GottfredsonStudyBehaviorLaw1979}, and \cite{KlingerNegotiatingOrderPatrol1997a}, refuses the idea that victims' characteristics can explain homicide clearance, arguing instead that alternative factors like weapon type or contextual circumstances influence the odds that a homicide will be solved. 

In spite of decades of theoretical speculation and empirical analyses, however, no universal consensus exists on the drivers of homicide clearances, leaving many questions still open - or even unasked. 

In light of this, the present work aims at further advancing scholarly knowledge on cleared homicides in the United States by focusing on national and state-wide patterns, unraveling the potential of explainable machine learning for studying such problem. This article expands the extant literature in three ways. 

First, the analysis will rely on the Murder Accountability Project (MAP) dataset, which is to date the most complete open-access dataset for studying homicide in the United States \citep{HargroveMurderAccountabilityProject2019}. The MAP dataset contains information on more than 774,000 homicides gathered from the FBI Supplementary Homicide Report enriched with information on additional 30,595 homicides that were not reported to the Uniform Crime Report (UCR) program, obtained through FOIA requests. 

Second, the work will  rely on a predictive machine learning approach, further diversifying the traditional quantitative set of techniques characterizing the literature on homicide clearances.

Third, in addition to the purely predictive component of the analysis, this work will frame national- and state-level results through the use of SHAP \citep{LundbergUnifiedApproachInterpreting2017}, an explainable machine learning framework, to show the promises that interpretable algorithmic approaches can offer to criminology.

The analytical framework will be divided into two main parts, the national and the state-level ones, each split into two additional sub-components: the predictive and the explainable machine learning ones. Code for the replication of the analyses are made available to encourage reproducibility and open science in criminology and crime research. 

The results of the study demonstrate the promises of machine learning, and particularly of XGBoost \citep{ChenXGBoostScalableTree2016}, a tree-based ensemble method, in correctly identifying cleared homicides and will offer confirmatory and new insights on the factors correlating with solved cases. However, state-level analyses also highlight the substantial variability in predictive performance, signaling that, although promising, there is no "one size fits all" machine learning approach for this specific application. Furthermore, state-level explainable models will also provide evidence on the heterogeneity of factors positively and negatively correlating with homicide clearance. This heterogeneity adds to the extant literature by showing that no single set of variables can stably and universally drive homicide clearance across the United States, reinforcing the necessity to avoid mutually exclusive theoretical explanations, as well as suggesting tailored interventions to improve police performance. 

The remainder of the paper proceeds as follows. The Background section will provide an overview of main perspectives and findings in research on homicide clearance, reflecting on possible theoretical paths forward. 

The Materials and Methods section will first describe the dataset and variables used in this study, and detail the feature selection and engineering phases for constructing the predictive models. Second, it will introduce the nine different algorithms tested and the SHAP interpretability framework. 

The Results section will be divided into two subsections, as anticipated, one on national-level analyses and one on state-level models, both presenting predictive as well as SHAP outcomes for algorithmic explainability. 

Finally, the Discussion and Conclusions section will synthesize the study's main results, reflect on its limitations and potential ways forward.

\noindent

\section{Background: Two Perspectives}\label{Back}

Homicide has been historically the focus of continuous attention in research and policy, given its seriousness, and the impact it has on the lives of offenders, victims, and communities. Homicides, in fact, do not only directly target and fatally harm victims, but also indirectly influence people that are close to them or that are part of the same community.  Given the central role that homicide acquired in criminology, the phenomenon has been scrutinized under several lenses. To name a few, researchers have analyzed its impact on criminal careers \citep{DeLisiUnpredictabilityMurderJuvenile2016}, relationship with socio-economic conditions \citep{ElgarIncomeinequalitytrust2011}, economic costs for the society \citep{DeLisiMurdernumbersmonetary2010}, link with social networks and community context \citep{PapachristosNetworkExposureHomicide2014}, and effect on life expectancy in communities with a high prevalence of violence \citep{RedelingsYearsYourLife2010}.

The overarching impact of homicide on society imposes the need for proper law enforcement response, which can be proxied by effectiveness in case clearance. Unsolved cases, in fact, can have tremendous multiplicative effects on indirect victims of lethal violence, including family, friends, colleagues, and neighbors \citep{ArmourExperiencesCovictimsHomicide2002, MastrocinqueStillLeftHere2015, BragaImprovingPoliceClearance2021}. Failing to solve a homicide further erodes the relationship between citizens and institutions, increasing the sense of insecurity and, in addition, creating the premises for retaliation, revenge or, in the case of serial offenders, new homicides \citep{RiedelDeclineArrestClearances1999,RegoecziUnclearedHomicidesCanada2000,BragaImprovingPoliceClearance2021}. 

Despite the relevance of the issue, the United States have witnessed a significant decline in arrest rates in the last decades \citep{OuseyKnowUnknownDecline2010, CouncilonCriminalJusticeHomicideTrendsWhat2021}, with some variation and few exceptions (see for instance \cite{Straussbetternabbingmurderers2017}). This trend has recently fostered the attention of both media and researchers, somehow responding to previous calls made by \cite{WellfordAnalysisVariablesAffecting1999} and  \cite{RiedelHomicideArrestClearances2008a} to devote more attention to the problem.  

Over the years two distinct theoretical perspectives were developed to explain why certain homicides are solved while others are not, in the attempt to provide evidence-based answers on this topic. The first perspective, which originates with Black's theory of law \citep{BlackBehaviorLaw1976}, is generally labeled as "discretionary" (or "extra-legal"); the second perspective is instead the "nondiscretionary" (or "solvability") one. 

On the one hand, the discretionary perspective implies that homicide investigators and detectives have some preferences towards certain types of victims. These preferences, in turn, are reflected in different degrees of commitment and amount of resources devoted to investigating homicides. 
This means that there are categories of victims for which higher efforts are deployed and categories of lower social status for which the vigor put into investigations is significantly lower, producing lower clearance rates. \cite{BlackBehaviorLaw1976} particularly argued that homicides with female, younger and non-white victims have lower odds of being cleared compared to homicides targeting male, older, white individuals.
After empirical scrutiny, however, findings turned out to be extremely heterogeneous across studies (and settings). Concerning gender, several works have reported that female and younger victims (especially children) are generally more prone to be part of homicides that are cleared by law enforcement \citep{AlderdenPredictingHomicideClearances2007a, Leevaluelifedeath2005, RegoecziClearingMurdersIt2008}. Yet, other studies found no effect, or even opposite direction of the relationship \citep{WolfgangPatternsCriminalHomicide1958, AddingtonUsingNationalIncidentBased2006, PuckettFactorsAffectingHomicide2003}. Similarly, the hypothesis that clearance rates hold a racial or ethnic component has found varying support in the literature so far. Some studies highlighted variations based on the racial background of the victims, particularly signaling higher odds of clearance for homicide involving White victims or lower odds for events with Black or Hispanic victims \citep{RegoecziUnclearedHomicidesCanada2000, Leevaluelifedeath2005, AlderdenPredictingHomicideClearances2007a, RegoecziClearingMurdersIt2008}, while others did not \citep{RiedelMurderClearancesMissing1996,AddingtonUsingNationalIncidentBased2006}.
Other categories that, according to the discretionary perspective, face the risk of being devalued in the investigative process are individuals with prior criminal histories and, more in general, those who live in urban areas with low socio-economic status, education levels, and social organization \citep{BlackBehaviorLaw1976}. Concerning the spatial distribution of homicides, for instance, recent evidence supported the idea that environmental features surrounding a homicide, e.g., physical environments and situational contexts \citep{KennedyEnvironmentalFactorsInfluencing2021}, as well as neighborhood characteristics - including their level of economic disadvantage - matter in clearing homicides \citep{MancikNeighborhoodContextHomicide2018a}.

On the other hand, the nondiscretionary perspective posits that the level of commitment and effort from law enforcement agencies is equal for all victims, regardless of their status, because of the inherent seriousness of the crime itself and the external pressure that society exerts on institutions for clearing all cases \citep{WolfgangPatternsCriminalHomicide1958, GottfredsonStudyBehaviorLaw1979}. Therefore, champions of this perspective theorize that different odds of solving a crime are mostly caused by factors like the context in which the event occurs, the weapon used, the location of the body, and area population surrounding the event. Concerning circumstances, for instance, homicides involving other felonies or gang activities, are less likely to be cleared \citep{CardarelliUnclearedHomicidesUnited1994,LitwinMultilevelMultivariateAnalysis2004, RobertsPredictorsHomicideClearance2007}. Conversely, homicides occurring between individuals knowing each other for motives as personal revenge, money-related arguments, or intimate affairs are much easier to solve, compared to those involving strangers \citep{WellfordAnalysisVariablesAffecting1999, RegoecziUnclearedHomicidesCanada2000}. 
Regarding weapons, homicides perpetrated with firearms or other tools that generally do not imply physical fight or contact are less likely to be solved, if compared against different types of weapons or modi operandi, such as the use of knives or strangulation \citep{PuckettFactorsAffectingHomicide2003, AddingtonUsingNationalIncidentBased2006}. Physical contact, in fact, leads in general to a greater amount of forensic evidence that can be used by law enforcement in their investigations, although findings regarding the role of forensic evidence on homicide clearance are also controversial in the literature \citep{Baskininfluenceforensicevidence2010, McEwenForensicEvidenceHomicide2015}, possibly due to the small samples and the prevalent use of city-level case-studies. Body location is considered important for investigating a murder as well, given that victims found in apartments or homes are deemed to be easier to clear. Finally, according to \cite{WolfgangPatternsCriminalHomicide1958}, area population is also a significant predictor, as homicides occurring in highly populated areas may offer perpetrators the possibility to escape or hide in anonymity, compared to communities with fewer people (and fewer suspects) with higher informal social control. \color{black} More recently, a line of empirical research further enriched the non-discretionary perspective by recommending to also consider law enforcement resources, training and organizational practices as potential correlates for clearing homicides \citep{McEwenForensicEvidenceHomicide2015, CarterEffectivePoliceHomicide2016, WellfordClearinghomicides2019, PizarroImpactInvestigationStrategies2020}. A study by \cite{BragaCanHomicideDetectives2018}, for instance, indicates that clearance rates can be improved through corresponding increases of resources dedicated to homicide detectives. Similarly, improving management resources has a positive effect on investigative activities, and the use of advanced analytical tools can also aid efforts in solving homicides: in this sense, \cite{WellfordClearinghomicides2019}  and colleagues set forth a series of recommendations based on an analysis of the association between investigative effort, case features, organizational factors, and homicide clearance. These invoke the role of resources applied to investigations and modalities through which work is organized. \color{black}

Works on homicide clearances have provided both clear, well-defined patterns, as well as contradictory evidence that do not allow to rule out completely either of the two perspectives reviewed above.

Leaving aside possible issues with research methodologies and designs, and inherent biases that may be hidden in the available data, this mix of stable outcomes and contradictory findings, may suggest that reducing the matter to a single theoretical perspective could be limiting for the complexity of the subject. This intuition is certainly not new: partially in line with this view, for instance, \cite{RegoecziHomicideInvestigationsContext2020} proposed a new approach based on Conjunctive Analysis of Case Configurations (CACC) for considering factors combining victims and event attributes to move beyond a complete blind separation of the two theoretical perspectives.

The multifaceted nature of homicide can hardly be framed monodimensionally: such monodimensionality leads to overlooking possible heterogeneity emerging across historical periods, jurisdictions, and organizations, and even hidden subpatterns at play when information on offenders, victims, and contexts are linked. This hypothesis of theoretical non-exclusivity applies to the United States and even more to international contexts, where homicides are less (or more) prevalent and society is constructed and organized differently. 

The present study tries to condense these two perspectives together by considering the use of variables related to discretionary and nondiscretionary prepositions and attempts to contribute to research on homicide clearance using an explainable machine learning approach. This approach acts on two geographical levels: a national one, to derive macro-level patterns, and a state-level one, to test for the presence of more localized dynamics that might provide support for the refusal of a single, mutually exclusive theoretical perspective. 

\color{black} The national and state-level focuses are motivated by the opportunity to gather knowledge on the possible heterogeneous patterns emerging when moving from a macro to a meso perspective, highlighting dynamics that are differentiated across state jurisdictions. Another alternative pathway would have been to concentrate on city-level dynamics given that homicide is a phenomenon that often falls within the jurisdiction of individual municipal agencies, yet this work only includes national and state-level focuses for three reasons. First, this is the first study allowing to compare all states, ensuring complete representativity of the entire country. Second, and related to the first point, the clustering dynamics of homicide in the United States and the resulting sparsity for many areas of the countries would have forced me to concentrate on a selected sample of major cities to carry out meaningful predictive models, thus losing the possibility to offer a comprehensive picture of the situation in the entire country. Third, analyzing homicide clearance patterns at the city-level without considering a state perspective would have posed significant problems in the interpretability and usability of the results, given the lack of an intermediate geographical focus. \color{black}

\section{Materials and Methods}

\subsection{Data}
\subsubsection{Overview: the Murder Accountability Project}
This paper exploits data on homicides committed in the United States from 1976 to 2019 retrieved from the Murder Accountability Project (MAP) website \citep{HargroveMurderAccountabilityProject2019}. The MAP is a non-profit organization committed to disseminating information related to homicides in the United States and improving the accountability of law enforcement institutions involved in their investigation. 

Specifically, the data utilized here include information on  737,836 homicide events and 774,156 homicide victims regularly reported in the well-known Supplementary Homicide Reports (SHR) published by the FBI Uniform Crime Report (UCR) program. Additionally, it also includes information on 30,595 events (and as many victims)  that were not reported in the SHR and were obtained by MAP through Freedom of Information Act (FOIA) requests (a distribution of cases gathered through FOIA is visualized in Figure \ref{fig:map}). The resulting is a victim-based dataset and not an incident-based one, meaning that each observation refers to a victim murdered in an event. Each observation will be labeled as a "homicide" throughout the work, although single events are lower than the number of homicides given the presence of cases in which offenders targeted multiple victims.

SHR have been largely used in the criminological literature to study homicides in the United States \citep{FoxHomicideElderlyResearch1991, BrowneGENDERINTIMACYLETHAL1993, Gallup-BlackTwentyYearsRural2005, RomanRaceJustifiableHomicide2013, ChanChoiceWeaponWeapon2016}. This also applies to homicide clearance in particular \citep{RegoecziUnclearedHomicidesCanada2000, OuseyKnowUnknownDecline2010, MancikHomicideclearancespre2019, ChalfinPoliceForceSize2020}. Yet, many have aptly pointed out the inherent problems associated with their characteristics \citep{BragaYouthHomicideBoston1999, BarberUnderestimatesunintentionalfirearm2002, LoftinAccuracySupplementaryHomicide2015}. The SHR are the most detailed among the various datasets in the UCR program, but the voluntary nature of data reporting by police agencies to the UCR program has been highlighted as one of the most important shortcomings concerning the reliability of the information reported in the SHR. 

This involves discrepancies between SHR and different data sources virtually measuring the same phenomenon, such as information collected by the Center for Disease Control and Prevention (CDC)  in the WONDER dataset and the Offenses Known and Clearances by Arrest data which are part of the UCR National Incident-Based Reporting System (NIBRS) program  \citep{KaplanChapterSupplementaryHomicide2020}. Comparative discrepancies translate into underreporting for SHR data, which hinders research (and policy) reliability of findings obtained by using such a source. 

The MAP initiative to obtain information on a substantial number of homicide cases that were never reported to the UCR program is thus a critical step forward in improving the quality and representativity of murder data in the United States. 

This article builds on this enriched version of the SHR dataset considering all the available events (filtering out observations with victim's age unknown led to the use of a total of 792,439 cases instead of the original 804,751). For the sake of clarity, throughout this work, I will refer to the MAP dataset as the one used to carry out the present study.

\subsubsection{Feature Selection and Engineering}
Besides the ID, which tracks each event, the MAP dataset contains a total of 30 different variables. It is worth noting that, compared to the traditional SHR datasets, when more than one perpetrator has participated in a given murder, information on the first one only is reported, while the SHR dataset includes information on up to 11 perpetrators. Unfortunately, it was unfeasible to perform exact matching between the original SHR and the MAP dataset to enrich this latter further by also including information of multiple perpetrators. The MAP dataset will thus only report the number of perpetrators involved without offering further detailed information on their characteristics. This limitation, however, affects only a total of 2,425 homicides, accounting for 0.3\% of all cases included in the MAP dataset.

Conversely, when a murder involves more than one victim, the MAP dataset reports a dedicated observation indexed by an identical ID mapping the specific event, such that every victim is recorded as separated. 

In terms of employed variables, only a subset of the total 30 available has been used for the present work. Besides the ID, information on the year in which the murder occurred has been retained as well as information on the month, the type of homicide, the age and sex of the victim, the circumstance, the weapon used, the type of agency that led the investigation on the case, the number and race of the victims, and the number of offenders. Additionally, a variable mapping whether another murder was investigated by the same agency in the same month of the same year and in the same city was created ad hoc. These variables were all exploited to predict whether a homicide (which, again, refers to a homicide of a single specific victim rather than an event with multiple casualties) was solved or not. Table \ref{desc} summarizes the variables included in all the models. 

\paragraph{Decade} The \textit{Decade} variable has been included because the ratio of solved homicides 
has not been stable across states over the years, with significant differences in the period under consideration \citep{LitwinDynamicNatureHomicide2007, XuCharacteristicsHomicideEvents2008}. Therefore, the decade in which a homicide has occurred can provide us with information about the level of predictability of our outcome of interest. 

\paragraph{Month}The \textit{Month} variable was selected to investigate whether different patterns can be distinguished across different periods of the year, hypothesizing that there may be seasonal clearance dynamics correlated with extant seasonal patterns in the occurrence of homicides \citep{McDowallSeasonalVariationHomicide2015}. In other terms, if homicides follow a certain degree of seasonality, with varying frequency over the course of the year, law enforcement agencies may be impacted by such seasonality in terms of reduction (or increase) in resources, thus influencing the probability that homicide is solved.

\paragraph{Homicide Type} The \textit{Homicide type} variable can only take two different values: "Murder and non-negligent manslaughter" and "Manslaughter by negligence". This latter category is associated in our data with unequal clearance rates (95.37\% of "manslaughter by negligence" cases have been solved, compared to only 70.18\% of "murder and non-negligent manslaughter" cases, which are by far the majority), and the rationale is thus to inherently distinguish between the two categories in the features space.

\paragraph{Age of Victim} The \textit{Age of Victim} variable, engineered through mutually exclusive 5-years spans, is included in the model following several findings in the extant literature on homicide clearance rates, which suggest how age is a significant predictor of clearance probability \citep{RegoecziUnclearedHomicidesCanada2000, PuckettFactorsAffectingHomicide2003, Leevaluelifedeath2005, AlderdenPredictingHomicideClearances2007a}.  \color{black} Particularly, this variable is theoretically situated in the nondiscretionary perspective.\color{black}

\paragraph{Sex of Victim} Similarly, the \textit{Sex of the Victim} variable has been long investigated in research on homicide clearance, with mixed findings regarding its correlation with the probability that homicide is cleared \citep{WolfgangPatternsCriminalHomicide1958, RiedelMurderClearancesMissing1996, PuckettFactorsAffectingHomicide2003, Leevaluelifedeath2005,AlderdenPredictingHomicideClearances2007a, RegoecziClearingMurdersIt2008}. \color{black} Specifically, the variable is linked to discretionary perspectives on homicide clearance \color{black} and was modeled through one-hot encoding leading to three levels, corresponding to female victims, male victims and unknown sex victims.

\paragraph{Race of Victim} \color{black} The focus on the race of the victim has been central in the development of the discretionary framework in homicide clearance studies. \color{black} Despite some works demonstrating variations across racial backgrounds, extant scholarship has not found universal consistency in the hypothesis that race correlates with homicide clearance \citep{CardarelliUnclearedHomicidesUnited1994, RiedelMurderClearancesMissing1996, PuckettFactorsAffectingHomicide2003, Leevaluelifedeath2005,AddingtonUsingNationalIncidentBased2006, AlderdenPredictingHomicideClearances2007a, RegoecziClearingMurdersIt2008, RobertsHispanicVictimsHomicide2011}. In light of this, \textit{Race of the Victim} is used as a feature to further investigate whether being part of a certain racial group correlates with solved (or unsolved) homicides. 

\paragraph{Number of Victims} The \textit{Number of Victims} is modeled in count terms and maps the extent to which a given event was associated with multiple murders. The underlying hypothesis motivating the inclusion of the variable is that more victims may imply higher odds of errors for the perpetrator or possibly more investigative leads for law enforcement, hence increasing the probability of homicide clearance. This relationship has been only marginally investigated in the previous literature \citep{Leevaluelifedeath2005, AddingtonHotvsCold2007, SturupUnsolvedhomicidesSweden2015}.

\paragraph{Circumstance} The \textit{Circumstance} variable provides information on the context in which the homicide occurred and \color{black} is traditionally the focus of the nondiscretionary theoretical perspective \color{black} \citep{WolfgangPatternsCriminalHomicide1958, LitwinMultilevelMultivariateAnalysis2004,LitwinDynamicNatureHomicide2007,AlderdenPredictingHomicideClearances2007a, LundmanExplanationsHomicideClearances2012}. \color{black} Including the "Circumstance" variable is particularly important because it allows to distinguish homicides besides information directly describing victims and offenders, adding a layer of contextual information that provides evidence on the motive or cause of the event. Homicides in the datasets are linked to a total of thirty-two different circumstances. The most frequent ones are "circumstances undetermined" (26.42\% of the cases), indicating homicides occurred in contexts were the motive or situation could not be ascertained, and "other arguments" (25.9\%), describing homicides happened in contexts like quarrels, interpersonal revenge, abuse or insult. The least frequent are homicide by abortion (0.001\%) and homicides resulting from gun-cleaning (0.01\%). \color{black}  One-hot encoding has been used to create dummy variables mapping all different possible circumstance levels.

\paragraph{Weapon Used} Information on \textit{Weapon Used} describes the type of weapon with which the victim was murdered. \color{black} Theoretically, it belongs to the nondiscretionary tradition \color{black} and, similarly to other features, numerous studies have already highlighted its impact in determining homicide clearances \citep{PuckettFactorsAffectingHomicide2003,AddingtonHotvsCold2007, RobertsPredictorsHomicideClearance2007}. \color{black} The dataset includes seventeen types of different weapons. The most frequent ones are "Handgun" (49\%), including pistol and revolvers, and "Knife or cutting instrument" (14.64\%). \color{black} This feature as well was enginereed via one-hot-encoding.

\paragraph{Number of Offenders} The variable \textit{Number of Offenders} has been modeled as a count variable measuring the number of individuals who took part in an event. The rationale for including the variable is that, similarly to the Number of Victims variable, homicides perpetrated in co-offending may be correlated with higher clearance probabilities due to higher odds of errors or defection. In line with this speculation, for instance, previous research indicated that multiple homicide offenders that act in teams have significantly shorter careers than those who act solo \citep{CampedelliSurvivalRecidivisticRevealing2021}, hence possibly implicating different odds of clearance for homicides characterized by solo versus multiple offenders. 

\paragraph{Monthly Overlap} The \textit{Monthly Overlap} variable has been created ad hoc for this study: it is encoded as a dummy variable, with 1 indicating that the same agency, in the same city, State, and month was already investigating another homicide. \color{black} Theoretically, the variable is in line with the nondiscretionary perspective, and particularly \color{black} with the hypothesis that police activity (and particularly resources, training, preparedness) plays a role in clearance rates \citep{KeelExploratoryAnalysisFactors2009, McEwenForensicEvidenceHomicide2015, WellfordClearinghomicides2019}. The rationale is to evaluate whether working under pressure may have a relationship with homicide clearance. Its importance should be higher for small agencies and small towns or cities rather than law enforcement agencies in cities like New York or Chicago, where resources are higher, and homicides occur much more frequently. 

\paragraph{Type of Agency} The last set of features is derived from the \textit{Type of Agency} variable, which was modeled using dummy variables mapping all agencies represented in the MAP dataset, \color{black} and adds to the nondiscretionary empirical line of inquiry that focus on police performance from an organizational perspective. \color{black} Its inclusion is motivated by the interest in evaluating whether variations exist for different agencies, to implicitly test if different human and investigative resources correlate with different odds in clearing homicides. \color{black} Specifically, seven categories of agencies are represented in the MAP dataset. Municipal Police is the most frequently observed (77.58\%), followed by Sheriff police (16.90\%). The least frequent is Tribal police (0.04\%).\color{black}

\paragraph{Target variable: Solved?} Finally, the \textit{Solved?} target variable is the outcome of interest and maps whether a given homicide was solved or not. The variable is created by the MAP exploiting information on the offender's sex: when it is unknown, the homicide is assumed to be unsolved. \color{black} The offender's sex variable is then cross-checked with other offender-related variables by MAP collaborators to ensure reliability of the criterion. These variables - which are not included in the models - are "offender's sex" and "situation" which maps whether the homicide involved solo, multiple or an unknown number of offenders. \color{black}  A similar strategy (using another variable) was carried out in \cite{RegoecziUnclearedHomicidesCanada2000}. There is a total of 529 cases all occurred in either Alabama or Florida that are labeled as solved but do not include information on offender's sex (0.065\% of total cases). They are likely cases in which information on the offender was not correctly reported. Since such information on offender's sex is not included in the models and the fact that a homicide is solved is expected to be less prone to errors compared to offenders-related variables, these homicides have been kept in the analyses. Section \ref{wp} in the Supplementary Materials provides additional analyses verifying the robustness of the target variable by comparing model outcomes with data obtained from \cite{TheWashingtonPostHowPostmapped2021}, and particularly from the "Murder with Impunity" project \citep{WashingtonPostMurderImpunity2018}.

A total of 236,692 homicides are labeled as unsolved in the MAP dataset, amounting to 29.41\% of total homicides registered in the United States from 1976 to 2019.\footnote{For further general reference, it should be noted that the FBI considers a homicide as solved either by arrest or exceptional means. For an homicide to be cleared by arrest,  at least one person has been arrested, charged with the commission of the offense, and turned over to court for prosecution. In cases cleared by exceptional means, instead, elements upon which law enforcement does not have control prevent the agency from arresting and charging the offender. Four conditions have to be met in this regard: the agency must have identified the offender, gathered enough evidence to support an arrest and consequent charge and prosecution, identified the offender's location, and encountered exceptional circumstances that prohibit the agency from arresting. One of such causes could be death or suicide.}

It is worth noting that other variables available in the MAP dataset that in principle may be important for analyzing characteristics associated with homicides have been excluded either for their low reliability (such as the ethnicity of the victim and the offender, which are seldom reported by agencies) or because they almost perfectly correlate with uncleared homicides (such as the variable describing the relationship between the victim and the offender, which in 92.6\% of the unsolved cases maps into a "non-determined" relationship between the two). 

\begin{figure}[hbt!]
    \centering
    \includegraphics[scale=0.42]{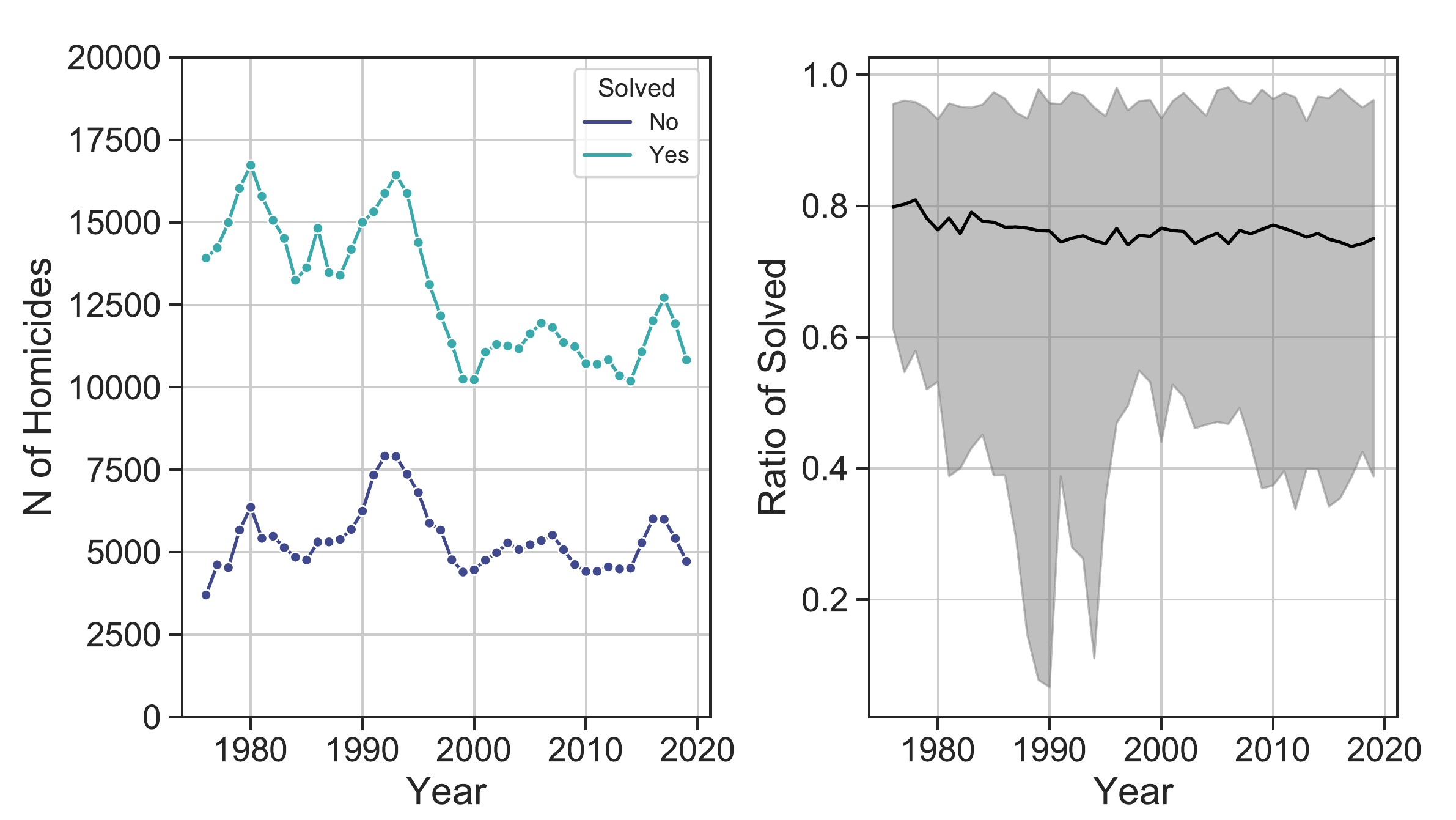}
    \caption{Yearly trends in solved and unsolved homicides at the national level (left) and trend in the state-level ratio of solved homicides out of total homicides, with the black line indicating average ratio for the United States, the upper gray area describing maximum yearly ratio and the lower gray area showing minimum yearly ratio (right).}
    \label{fig:ts}
\end{figure}

\color{black} The yearly trends of solved and unsolved homicides and state-level ratio of solved homicides out of total homicides are displayed in Figure \ref{fig:ts}. It is worth noting the wide variability characterizing ratios at the state level, with large standard deviations demonstrating very different performances across states (and years). While it is beyond the scope of this paper to understand what drives these stark differences, it is nonetheless important to underline the finding and possibly stimulate reflections on its possible multifaceted causes. \color{black} 
\begin{table}[!hbt]
\setlength{\tabcolsep}{1.3pt}
\footnotesize
\begin{tabular}{lll}
\cline{1-2}
\textbf{Variable} & \textbf{Format} &  \\ \cline{1-2}
ID & Alphanumeric. Indexes each case &  \\
Decade & Numeric. Processed through one-hot encoding. &  \\
Month & String. Processed through one-hot encoding. &  \\
Homicide Type & String. Processed through one-hot encoding &  \\
Age of the Victim & Numeric. Multiple categories created (5 years span). Processed through one-hot encoding. &  \\
Sex of the Victim & String. Processed through one-hot encoding. &  \\
Race of the Victim & String. Processed through one-hot encoding. &  \\
Number of Victims & Numeric. Number of total victims involved in the same event. &  \\
Circumstance & String. Processed through one-hot encoding. &  \\
Weapon Used & String. Processed through one-hot encoding. &  \\
Number of Offenders & Numeric. Number of total offenders involved in the same event. &  \\
Monthly Overlap & Numeric. Maps whether same agency in same (month;year;State) was investigating another case. &  \\
Type of Agency & String. Processed through one-hot encoding. &  \\
Solved? & String. Target Variable. Processed through one-hot encoding. &  \\ \cline{1-2}
\end{tabular}
\caption{Summary of Variables from the MAP Dataset Used in the Predictive Models.}
\label{desc}
\end{table}

%% until here
\subsection{Methodology}
\subsubsection{Analytical Strategy}
The study is primarily centered around two separate (and yet interrelated) tasks. First, predicting solved and unsolved homicides at the national and state levels to test the performance of machine learning approaches in correctly classifying events based on a set of features that combine information on victims, offenders, context (e.g., weapon, circumstances), and law enforcement. Second, by maintaining the focus on both national and state dimensions, explaining the nature of the obtained predictions to disentangle what factors help discriminate solved and unsolved homicides, further investigating possible variations and heterogeneity of correlates and drivers.

Both tasks rely on explainable machine learning as the methodological backbone. Machine learning is increasingly gaining momentum in criminology and criminal justice  \citep{BrennanEmergenceMachineLearning2013c, BerkAlgorithmiccriminology2013d, CampedelliWherearewe2021}. In response to lively debates regarding the "black box" nature of predictions and recommendations offered by machine learning algorithms in high-stakes applications, including those related to policing, criminal justice, and healthcare, scholars in Artificial Intelligence and computer science have recently proposed several approaches to increase model interpretability, fairness and accountability \citep{HolzingerWhatweneed2017, RudinStopexplainingblack2019a, GunningXAIExplainableartificial2019, AngelovExplainableartificialintelligence2021}. This work leverages such advances, combining predictions with explainability on both analytical levels.

\paragraph{Prediction Tasks} 
Concerning the first task, nine different algorithmic approaches have been tested, all characterized by different combinations of relevant hyperparameters. These are Ridge Regression (Ridge), Lasso Regression, (Lasso) Elastic Net Regression (EN), Decision Trees (DT), Random Forests (RF), Gradient Boosting Machines (GBM), Linear Support Vectors Machines (SVM), XGBoost (XGB), and Linear Discriminant Analysis (LDA). National-level performance across algorithms is assessed through grid search of relevant hyperparameters, to fine-tune the performance of the different approaches given that in many cases distinct values for distinct hyperparameters can lead to significant differences in model outcomes. Details about the models and the related hyperparameters are available in the Supplementary Materials section \ref{algoapp}.  After the predictive step at the national level, the algorithm that was found to perform best is used to learn state-level predictions.

For both national- and state-level models, the datasets have been split into train and test sets, using 70\% of the data points for the training phase and the remaining 30\% for the actual prediction phase. \color{black} To avoid issues arising with the presence of possible temporally-dependent patterns and covariate shift, I proceeded to random shuffling of observations before splitting data into train and test set. \color{black}

Model evaluation has been carried out through 5-fold stratified cross-validation. Cross-validation is a popular technique designed to obtain stable and reliable predictions, aiming to improve generalization and reduce overfitting. K-fold validation specifically works by picking a number of data folds (5 in this case), selecting k-1 folds as the training test, with the remaining fold used as the test set. The model is then validated on the training set used as test, and the procedure is repeated such that at the end, the model is validated using all folds as test sets, finally averaging performance across all folds. I have employed the stratified version of cross-validation, ensuring that each fold keeps the same proportion for the two classes in the target variable, given that the distribution of solved and unsolved homicides is not balanced both country-wise and state-wise.

To evaluate the performance of the models, I have used Balanced Accuracy and Precision. Balanced Accuracy is an alternative to ordinary accuracy, intended to work better in cases with imbalanced classes. In fact, overall accuracy offers unrealistic performance estimates in applications where one class appears more often than the other. 

Precision is a metric that seeks to quantify the extent to which a classifier avoids labeling negative (i.e., unsolved) homicides as positive (i.e., solved). Given that solved homicides are the majority in the national and state-level datasets, the rationale behind this choice is to minimize false positives. A detailed description on how these two metrics are computed is available in the Supplementary Materials section \ref{perfo}. 
\paragraph{Model Explainability through SHAP} 
A second important component of this work relates to the assessment of model outcomes through explainable machine learning, and particularly SHAP, which stands for SHapley Additive exPlanations, a method designed to explain individual predictions \citep{LundbergUnifiedApproachInterpreting2017}.

SHAP originates from coalitional game theory \citep{ShapleyValueNPersonGames1952} and is also inspired by previous approaches for model interpretability \citep{StrumbeljExplainingpredictionmodels2014, DattaAlgorithmicTransparencyQuantitative2016, RibeiroWhyShouldTrust2016a}.The method departs from the idea that a prediction can be described as a payout in a game in which every feature represent a player: the fundamental goal is then to learn how to fairly distribute the payout among features based on their actual contribution. SHAP provides an estimate of how much a given feature has contributed to the prediction, compared to the average predicted value. In other terms, the SHAP value $\phi_i$ for a given feature $i$ is the contribution of that same feature to the payout (i.e., prediction), weighted and summed over all the possible feature value combinations: 

\begin{equation}
    \phi_i=\sum_{S\subset N\setminus \left\{ i \right\}}\frac{|S|!(|N|-|S|-1)!}{|N|!}(f_{S\cup \left\{ i \right\}}(x)-f_S(x))
\end{equation}

where $S$ represent a subset of all available features $F$, $S\cup \left\{ i \right\}$ is the model trained including the variable of interest $i$ and $f_S$ is another model trained without $i$, and $f_{S\cup \left\{ i \right\}}(x)-f_S(x)$ represents the difference between predictions for the two models. Given that the effect of leaving out a variable varies based on the features that are present in the model, the difference $f_{S\cup \left\{ i \right\}}(x)-f_S(x)$  is computed for all the potential subsets of features $S\subset N\setminus \left\{ i \right\}$. In this study I have employed TreeSHAP, an expansion of the original SHAP method designed for tree-based and ensemble methods that proposes a more efficient computational approach to estimate $f_S(x)$ \citep{LundbergExplainableAITrees2019}.

\begin{comment}
SHAP satisfies three fundamental properties that guarantee the fair allocation of payout for the importance of the contribution of each feature in a predictive model, namely \textit{local accuracy, missingness} and \textit{consistency}.
\end{comment} 
In general, SHAP holds two critical practical benefits: first, it enables global interpretability of models, quantifying the overall impact of features on model output; second, it offers local interpretability, such that each individual prediction can be scrutinized and evaluated, enhancing transparency.

\section{Results}

\subsection{Predicting Cleared Homicides at the National Level}

\subsubsection{National-level Predictive Accuracy}
\begin{figure}[!hbt]
\centering
\begin{subfigure}[b]{0.95\textwidth}
   \includegraphics[width=1\linewidth]{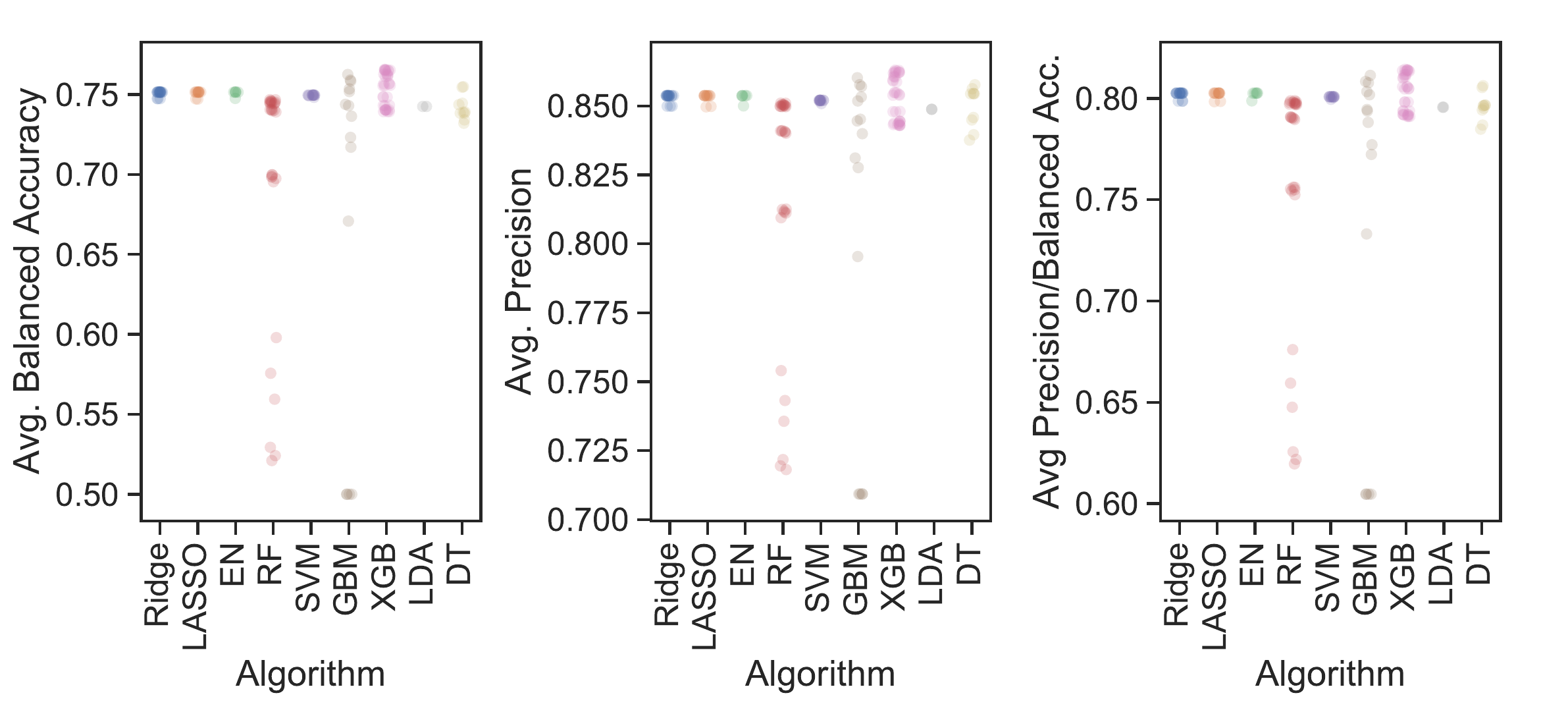}
   \caption{}
   \label{subfig:Ng1} 
\end{subfigure}

\begin{subfigure}[b]{0.55\textwidth}
   \includegraphics[width=1\linewidth]{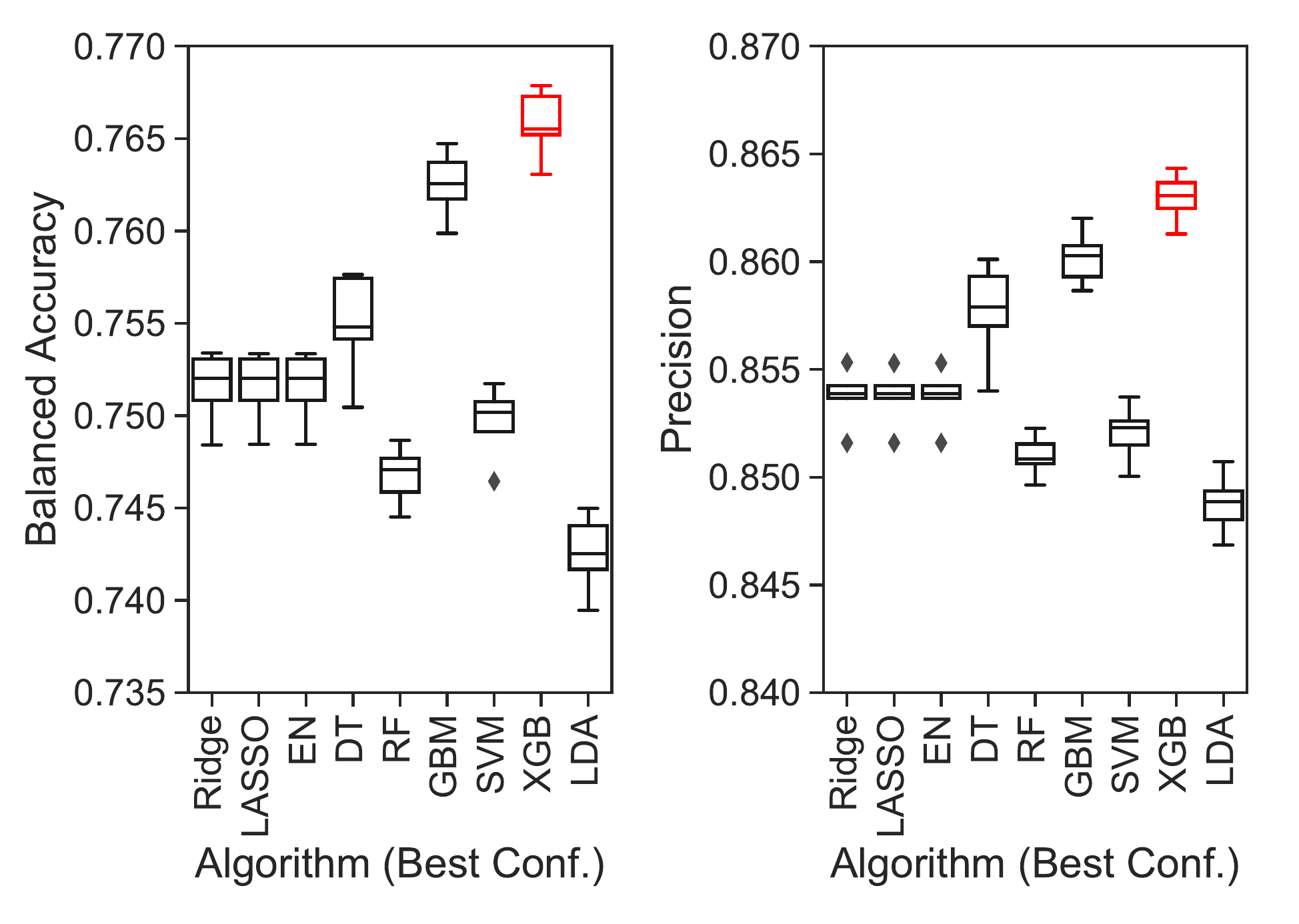}
   \caption{}
   \label{subfig:Ng2}
\end{subfigure}
\caption{(a): Distribution of algorithmic performance across average training Balanced Accuracy, average training Precision and average of average training Balanced Accuracy and average training Precision. XGBoost demonstrates to achieve superior performance compared to alternatives. (b): Distribution of training Balanced Accuracy and Precision for best configuration in each algorithm, data points indicate performance on single cross-validation split. Red indicates best performance.}
\label{fig:algo_perf}
\end{figure}

As already mentioned, a total of nine different algorithmic approaches have been tested through extensive grid search for optimization of hyperparameters. The distribution of the performance of each configuration for each algorithmic approach is displayed in Subfigure \ref{subfig:Ng1}. Each point represents the average performance of each cross-validated model (with cross-validation being 5-fold stratified cross-validation), and three measures of performance are offered to summarize the available information: besides average training Balanced Accuracy and average training Precision, the right subplot provides the average of the two. Two results emerge from Subfigure \ref{subfig:Ng1}: First, XGBoost, for more than one configuration, is the algorithmic approach that obtains the highest performance across all nine methods. Second, Ridge, LASSO, and Elastic Net demonstrate to be almost perfectly equal in their performances, which are considerably high and very stable across different configurations. Third, Random Forests and Gradient Boosting Machines manifest very large variations in their performance. Fourth, and more in general, except for some poorly performing outliers, the overall predictive power of the nine approaches is highly comparable. 

Subfigure \ref{subfig:Ng2} further highlights the alignment between the performances. The figure shows the comparison of the best configuration for all nine methods, with data about each cross-validation split visualized (information on the hyperparameters in each best configuration is reported in Table \ref{table: bconb}). XGBoost is, as anticipated, the algorithm with the highest performance. The mean average Balanced Accuracy is 0.765, with a standard deviation of 0.001, and the mean Precision is 0.865, with a standard deviation equal to 0.001. In terms of Balanced Accuracy, the percentage variation between the cross-validated version of the best configuration of XGBoost and its competitors ranges from +0.58\% (Gradient Boosting Machines) to +3.10\% (Linear Discriminant Analysis). In terms of Precision, the range is even more restricted: from +0.34\% (against Gradient Boosting Machines) to +1.64\% (LDA). In terms of raw classification terms and particularly missclassified events in the test set, this amounts to a difference of  523 fewer incorrectly classified cases compared to GBM and 4,156 compared to LDA.  
When considering the performance of the each best configuration on the test sets, results remain very similar, both in absolute and relative terms. 

In general, these results indicate that while there may be more than one possible candidate for achieving robust predictive results, not all machine learning methods are at the same level in terms of accuracy, showcasing the importance of grid search as a filtering approach in algorithmic comparisons.  

\begin{table}[!hbt]
\centering
\scriptsize
\setlength{\tabcolsep}{3pt}
\begin{tabular}{lclcccc}
\hline
\textbf{Algorithm} & \textbf{\begin{tabular}[c]{@{}c@{}}N Tested\\ Config.\end{tabular}} & \multicolumn{1}{l}{\textbf{\begin{tabular}[c]{@{}c@{}}Hyperparameters\\ (Best Config.)\end{tabular}}} & \textbf{\begin{tabular}[c]{@{}c@{}}CV Avg. Train.\\ Balanced\\ Accuracy\end{tabular}} & \textbf{\begin{tabular}[c]{@{}c@{}}CV Avg. Train\\ Precision\end{tabular}} & \textbf{\begin{tabular}[c]{@{}c@{}}Test\\ Balanced \\ Accuracy\end{tabular}} & \textbf{\begin{tabular}[c]{@{}c@{}}Test\\ Precision\end{tabular}} \\ \hline
Ridge & 18 & \textit{Solver: liblinear, C: 50} & \begin{tabular}[c]{@{}c@{}}0.752 \\ (0.002)\end{tabular} & \begin{tabular}[c]{@{}c@{}}0.854 \\ (0.001)\end{tabular} & 0.751 & 0.854 \\ \hline
LASSO & 12 & \textit{Solver: saga, C: 50} & \begin{tabular}[c]{@{}c@{}}0.752 \\ (0.002)\end{tabular} & \begin{tabular}[c]{@{}c@{}}0.854 \\ (0.001)\end{tabular} & 0.751 & 0.854 \\ \hline
Elastic Net & 6 & \textit{Solver: Saga, C: 50,  L1 ratio: 0.5} & \begin{tabular}[c]{@{}c@{}}0.752 \\ (0.002)\end{tabular} & \begin{tabular}[c]{@{}c@{}}0.854\\ (0.001)\end{tabular} & 0.751 & 0.854 \\ \hline
DT & 6 & \textit{Criterion: Gini, Max Tree Depth: 20} & \begin{tabular}[c]{@{}c@{}}0.747 \\ (0.003)\end{tabular} & \begin{tabular}[c]{@{}c@{}}0.851\\ (0.003)\end{tabular} & 0.756 & 0.858 \\ \hline
RF & 48 & \textit{\begin{tabular}[c]{@{}l@{}}Criterion: Gini, Max Tree Depth: 50,  \\ N of Estimators: 200\end{tabular}} & \begin{tabular}[c]{@{}c@{}}0.744 \\ (0.001)\end{tabular} & \begin{tabular}[c]{@{}c@{}}0.849\\ (0.000)\end{tabular} & 0.749 & 0.853 \\ \hline
GBM & 18 & \textit{Learning Rate: 0.5, N of Estimators: 200} & \begin{tabular}[c]{@{}c@{}}0.763 \\ (0.002)\end{tabular} & \begin{tabular}[c]{@{}c@{}}0.860\\ (0.001)\end{tabular} & 0.764 & 0.861 \\ \hline
SVM & 12 & \textit{Penalty: L2, C: 50} & \begin{tabular}[c]{@{}c@{}}0.750 \\ (0.002)\end{tabular} & \begin{tabular}[c]{@{}c@{}}0.852 \\ (0.001)\end{tabular} & 0.750 & 0.853 \\ \hline
XGBoost & 36 & \textit{\begin{tabular}[c]{@{}l@{}}Learning Rate: 0.5, N of Estimators: 200, \\ Gamma: 0\end{tabular}} & \textbf{\begin{tabular}[c]{@{}c@{}}0.766\\ (0.001)\end{tabular}} & \textbf{\begin{tabular}[c]{@{}c@{}}0.863 \\ (0.001)\end{tabular}} & \textbf{0.767 }& \textbf{0.863} \\ \hline
LDA & 6 & \textit{Shrinkage: Ledoit-Wolf, Solver: least square} & \begin{tabular}[c]{@{}c@{}}0.743 \\ (0.001)\end{tabular} & \begin{tabular}[c]{@{}c@{}}0.848 \\ (0.000)\end{tabular} & 0.744 & 0.850 \\ \hline
\end{tabular}
\caption{Summary of Performance for best hyperparameter configuration per each algorithmic approach. Bold figures indicate the best model across all.}
\label{table: bconb}
\end{table}

\subsubsection{National-level Predictive Explainability}

SHAP values have been computed on the test set using the same best-performing model among the nine tested algorithms, with its best configuration, namely XGBoost with 200 estimators and a learning rate of 0.5. SHAP values are here reported in terms of log odds, meaning that they can take negative and positive results to indicate their contribution to the final prediction outcome. 
\begin{comment}
Similar to the traditional interpretation of coefficients in logistic regression, a positive log-odds means that a given feature positively contributes to the prediction of a cleared homicide. In other words, if a feature has a positive log-odds, it means that a homicide being associated with that variable has a higher probability of being a solved one. Conversely, a negative log-odds indicates that a given variable has a negative impact on the prediction of a cleared homicide, meaning that it is negatively related to the outcome. Trivially, log odds equal to 0 have no role in modifying the model output magnitude. 
\end{comment}
For reference, SHAP also allows inspecting local predictions and explaining in detail the impact of each feature associated with a given event on the final predicted outcome. Two such examples are provided in the Supplementary Material, Figures \ref{fig:2figsA} and \ref{fig:2figsB}.

\begin{figure}[!hbt]
    \centering
    \includegraphics[scale=0.4]{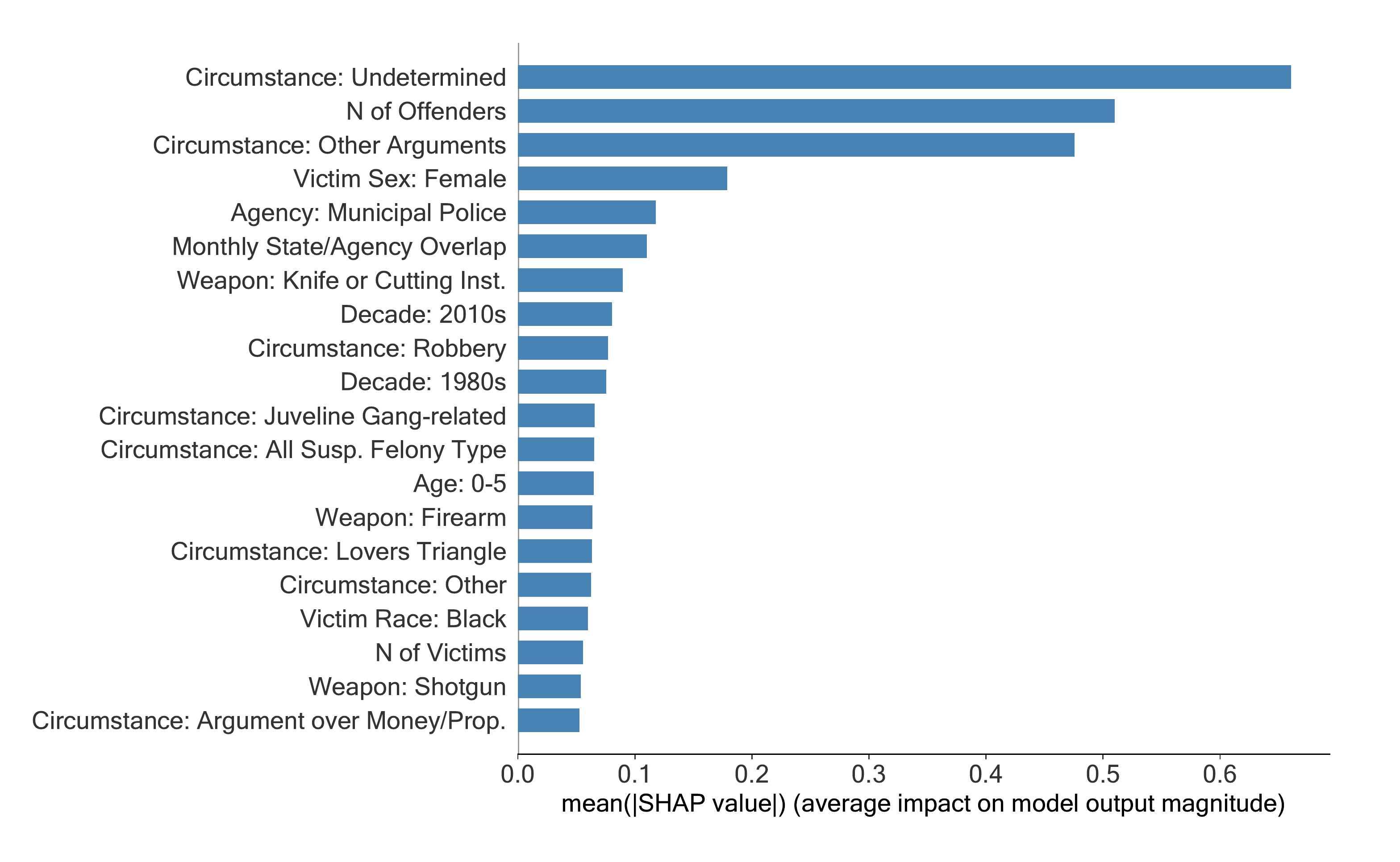}
    \caption{List of top 20 features with highest average SHAP value in absolute terms, mapping average impact on US model output magnitude.}
    \label{absshap}
\end{figure}

Figure \ref{absshap} reports the list of top 20 features in terms of their mean SHAP values in absolute terms in the test set, comprising a total of 237,732 homicides. This quantity assesses the overall average impact of a given feature on the outcome, i.e., its average impact on model output magnitude. Since SHAP values are calculated per each observation, it should be noted that the impact of a feature may be very high in one case, and very low in another case. This may be related to patterns existing in certain regions of the data, where opposite trends are present, or particular combination of values among different features. Absolute values are hence helpful as a first overall measure of impact. 

Figure \ref{absshap} shows that the feature "Circumstance: Undetermined" is by far the most impactful overall, followed by the "Number of Offenders", "Circumstance: Other Arguments" and "Victim Sex: Female". Features related to homicide circumstances are prevalent in the list of top 20 most impactful features: 7 out of 20 are associated with information on the context surrounding the event ("Circumstance: Undetermined", "Circumstance: Other Arguments", "Circumstance: Robbery", "Circumstance: Juvenile Gang-related", "Circumstance: All Suspected Felony Type", "Circumstance: Lovers Triangle", "Circumstance: Other", "Circumstance: Argument over Money or Property"). Another important group of features (4 out of 20) correspond to those related to characteristics of the victims: "Victim Sex", "Age: 0-5", "Victim Race: Black", "Number of Victims". Three variables are related to weapons ("Weapon: Knife or Cutting Instrument", "Weapon: Firearm", "Weapon: Shotgun"), two to agency-level information ("Monthly State/Agency Overlap" and "Agency: Municipal Police") and two concern temporal information about the decade in which a homicide has occurred ("Decade: 2010s" and "Decade: 1980s"). 

\begin{figure}[!hbt]
    \centering
    \includegraphics[scale=0.55]{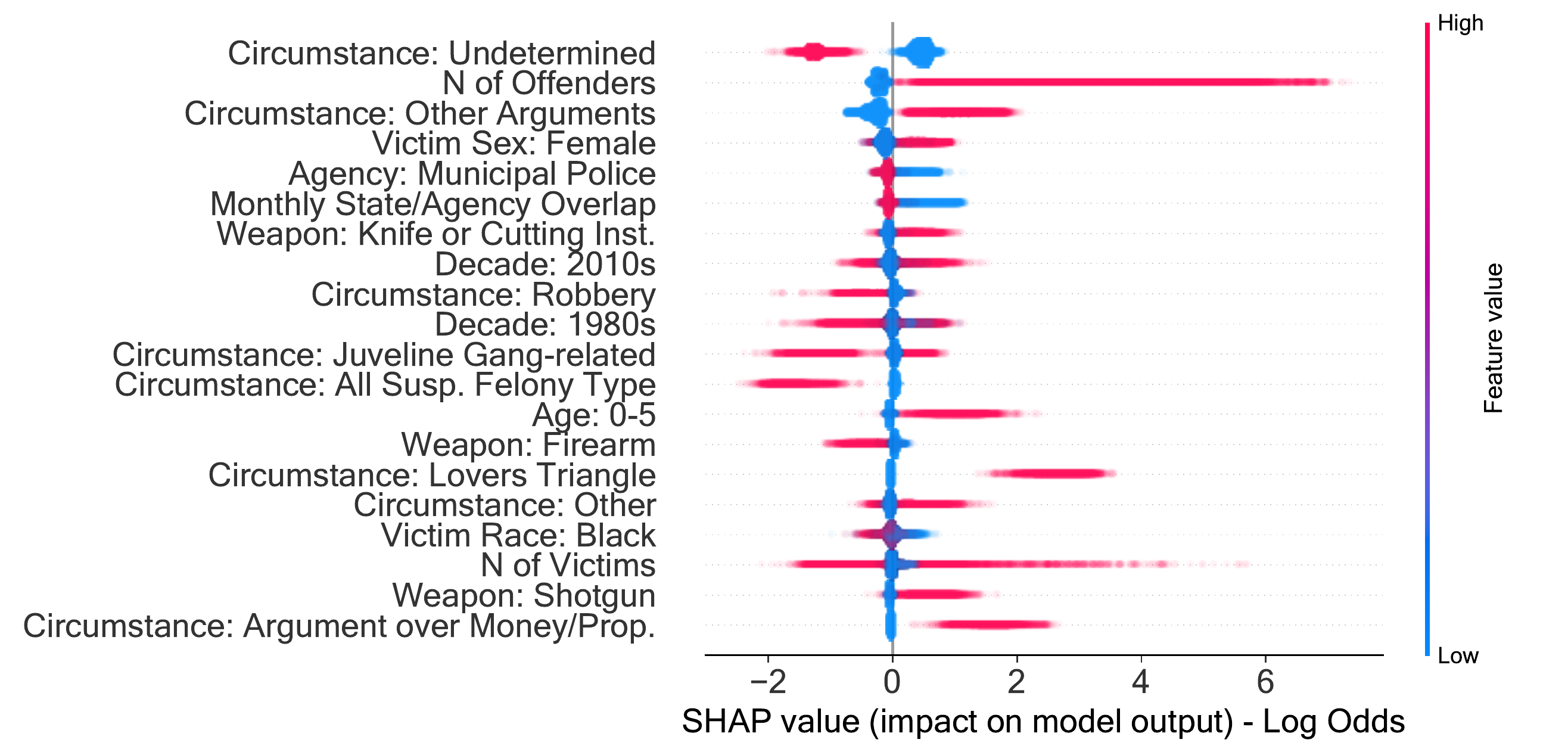}
    \caption{Distribution of SHAP values for the top 20 most impactful features at the national level. Each point represents an observation in the US test set. Points falling left of the zero threshold mean a negative relationship with cleared homicides, while points falling right indicate otherwise. Each point is colored based on the value of the feature: in the case of a binary variable, if the instance is equal to 1 for a particular observation, the point will be colored in red, blue otherwise. Nuanced colors will identify count variables, e.g., Number of Offenders. }
    \label{fig:shap_dist}
\end{figure}

Figure \ref{fig:shap_dist} translates the information synthesized in Figure \ref{absshap} into the distribution of the actual SHAP values for each of the 20 most impactful features in the national-level model. Each point in each row indicates an observation. As shown by the color bar, when a point is colored in red, the value of the feature associated with each point is low, when the point is blue, instead, the value is high. For instance, if an observation $i$ on the binary feature $k$ equals 1, the corresponding point will be blue, and red otherwise. The location of the point is also important: if the point falls left to the zero threshold, it means that the given feature $k$ for observation $i$ has a negative impact on the final prediction, expressed in log-odds.

On the contrary, if the point falls right, the impact is positive (or, in other words, the relationship of that instance with the prediction is positive). Points clustered around zero mean very low or no impact at all. The higher the distance to the zero threshold, the higher the impact (either negative or positive) on the model. 

The most impactful feature overall ("Circumstance: Undetermined") here indicates that when investigators are not able to reconstruct the context around the homicide, the odds that it will go uncleared are very high, as shown by the positive spread of the cloud of blue points falling right to the zero threshold. A similar relationship is found for "Circumstance: All Suspected Felony Type", mapping events in which there is a suspect that a possible felony occurred along with the homicide, although evidence or facts do not guarantee clear determination of circumstances.

Contrarily to undetermined circumstances, when a homicide occurs in the context of quarrels or interpersonal conflicts as abuse, insult or revenge, falling into the "Circumstance: Other Arguments" category, the probability that the homicide will be cleared is considerably high, with a neat distinction between feature values in the two regions around the zero threshold. This also applies to alternative circumstances, like those involving lovers' triangles ("Circumstance: Lovers Triangle") and arguments over money or property ("Circumstance: Argument over Money/Property"). In other cases, circumstances have opposite relationships with the outcome, or findings are mixed. Most features clearly separate negative and positive impact on model prediction, meaning that high or low feature values fall distinctly in one of the two regions. 

However, as anticipated in the previous paragraph commenting Figure \ref{absshap}, there may be cases in which the impact of a feature with the same value (e.g., $k=1$ on observations $i$ and $j$) are very different, and even of opposite sign. Homicides occurred in the context of a robbery ("Circumstance: Robbery"), falling into the residual "Circumstance: Other" category or those associated with juvenile-gang killings ("Circumstance: Juvenile Gang-related") are characterized by mixed outcomes. In the first case, if a homicide is associated with a robbery, the odds of solving it are lower, although as shown by the red points right to the zero threshold, there are cases in which robberies correlate with solved homicides. Mixed findings are even more evident in the "Circumstance: Other" case: this is probably due to the different ways in which agencies across the country label homicides circumstances, thus reflecting in the way in which this residual category is used. In terms of "Circumstance: Juvenile Gang-related", the mixed findings (i.e., very high or very low odds of solved homicides when an event belong to this category), may reveal localized subpatterns present in the data, possibly associated with different levels of investigative effectiveness in the context of gang violence.

Victim-level features provided both clear and mixed results as well. Concerning clear patterns, homicides involving female victims ("Victim Sex: Female") have considerably higher odds of being solved. This finding may be explained by the fact that women are often killed in contexts where circumstances are clear (e.g., intimate partner violence), and investigations are hence easier. 

Homicide involving children ("Age: 0-5"), and particularly those aged between 0 and 5 years, also have strongly positive odds of being cleared, and the reason may be the smaller circle of people that are part of a child's network, added to minimal mobility, reducing the probability of fatal encounters with strangers and, most importantly, raising the likelihood that perpetrators are in the context of the family.

Events involving Black victims ("Victim Race: Black") are also associated with lower odds of being cleared. While the general trend suggests that these events are more likely to go unsolved, the overlap of mixed findings around the zero threshold call for further analysis to disentangle the different mechanisms at play. 

Further analyses are also required concerning the role that the number of victims ("N of Victims") has on the odds of clearing a homicide. The presence of red points widely distributed both below and above zero, suggesting that there may exist differences (most likely across states) in the effectiveness through which these homicides are investigated. 

In terms of offender-related information, when more individuals participate in a homicide ("N of Offenders"), the odds that homicide is cleared sensibly increase, in line with the hypothesis stated in the Methodology subsection that more offenders imply a higher risk of leaving evidence, confessions, and errors. 

Weapons also reveal clear patterns: homicides perpetrated with the use of a knife or cutting instrument ("Weapon: Knife or Cutting Inst.") have considerably higher odds of being solved, although with very residual exceptions (i.e., the red points falling below the zero threshold). On the contrary, homicides with firearms or shotguns are negatively related to solved events. 

In terms of agency-related features, when an agency in a given city is investigating another homicide in the same month ("Monthly State/Agency Overlap"), the odds that the homicide in consideration will be solved decrease (although, again, with some residual exceptions). This might be explained by the fact that most small agencies do not have enough resources, training, or time to manage two murder investigations simultaneously. Additionally, homicides investigated by agents of municipal police agencies ("Agency: Municipal Police") have lower odds of clearance, with a similar distribution to the one displayed by the "Monthly State/Agency Overlap" feature.

Finally, mixed findings emerge for both the two decades included in the list ("Decade: 2010s" and "Decade: 1980s"). These outcomes possibly hide variations in clearance rates across agencies and, reasonably, states in the two decades. 

\subsection{Predicting Cleared Homicides at the State Level}

% up until here
\subsubsection{State-level Predictive Accuracy}
Given that XGBoost demonstrated to be the most powerful among the nine approaches tested for predicting cleared homicides at the national level, it was used as the approach for analyzing homicides and predicting solved instances also at the state level. Yet, grid search for hyperparameter selection has been performed for each state dataset via the same hyperparameters considered at the national level, except for the Gamma regularization term which was found to cause no meaningful variations in performance compared to non-regularized models. Grid search was performed because state-level datasets consistently vary in terms of observations as well as distribution of positive and negative examples (a graphic illustration can be appreciated in Supplementary Materials Figure \ref{fig:desc_cases}). Hence, tailored optimization was required to maximize prediction accuracy in the first place and obtain reliable prediction explanations in the second place. A total of $3\times4\times51=612$   models have been run  overall, given that 3 estimators (50, 10, 200) and 4 learning rates (0.01, 0.1, 0.2, 0.5) were evaluated  to obtain the 51 state-level optimized predictive models.

\begin{figure}[!hbt]
    \centering
    \includegraphics[scale=0.32]{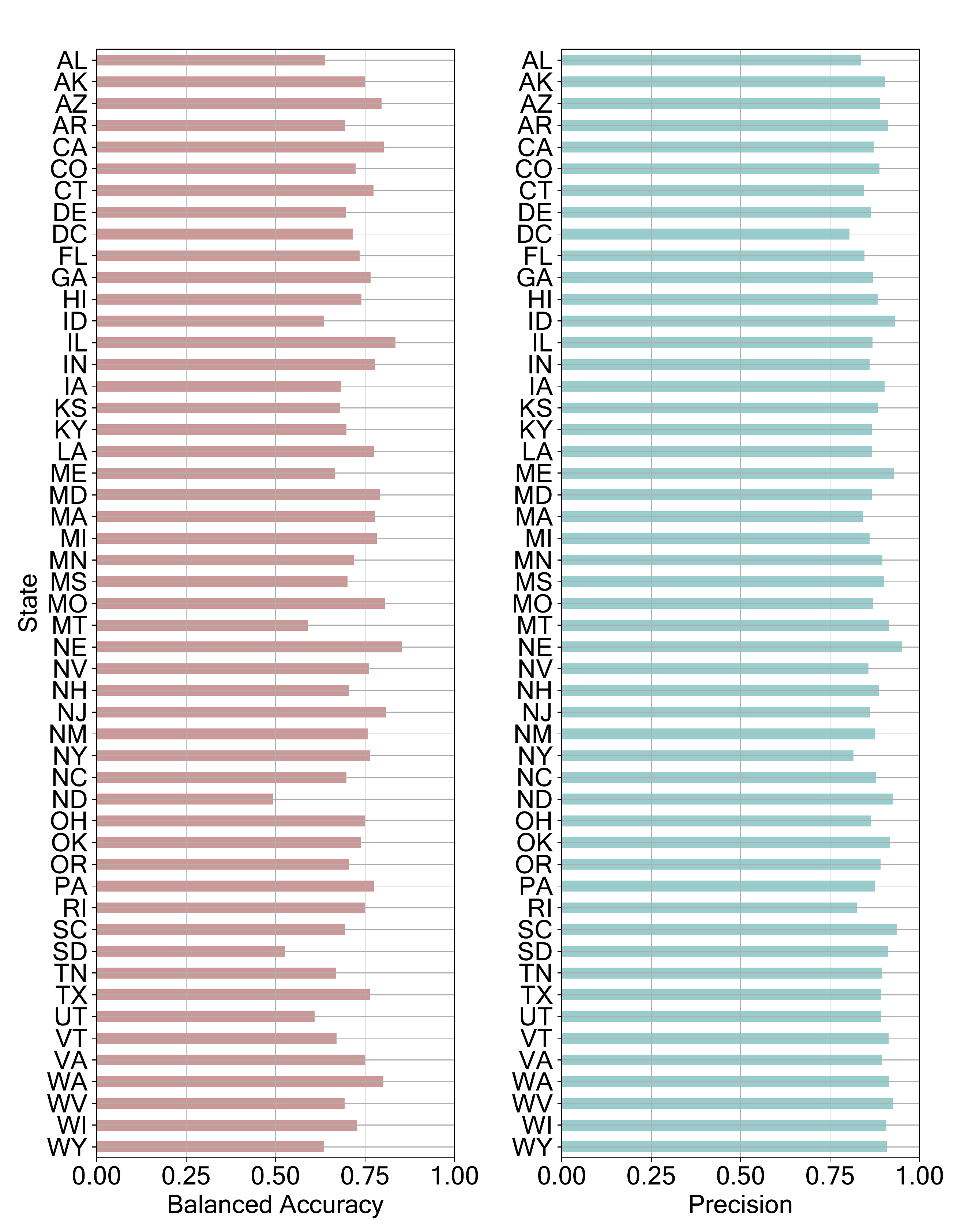}
    \caption{Test Balanced Accuracy and Test Precision for all states represented in the MAP dataset based on the best state-wise XGBoost hyperparameter configuration.}
    \label{fig:state-level-perf}
\end{figure}

Once configurations have been tested through cross-validation for all states, the best candidate for each was chosen and deployed. Figure \ref{fig:state-level-perf} shows the performance of the best XGBoost configuration in each state in terms of Balanced Accuracy and Precision on the test sets. Concerning Balanced Accuracy, Nebraska (0.853), Illinois (0.837),  New Jersey (0.809), Missouri (0.805), California (0.802), and Washington (0.801) are the states achieving the best predictive results. North Dakota (0.492), South Dakota (0.526),  and Montana (0.591) have instead the lowest Balanced Accuracy. The average Balanced Accuracy across states is 0.722, with a standard deviation of 0.07.

In terms of Precision, which assesses the ability of the algorithms to reduce false positives, Nebraska (0.951), South Carolina (0.936), Idaho (0.932), Maine (0.927), West Virginia (0.927), and North Dakota (0.925) obtain the highest scores. The lowest results are achieved by District of Columbia (0.804) and New York (0.815), which reach relatively high outcomes nonetheless. The average Precision across states is 0.884, with a standard deviation of 0.03.
Interestingly, North Dakota is among the lowest Balanced Accuracy scores, but among the highest in terms of Precision. This discrepancy can be due to the fact that Precision does not use false negatives and true negatives in its computation, while Balanced Accuracy does. The difference then can be explained by low ability of  XGBoost in North Dakota to classify negative examples correctly. This finding (and in general the Pearson's correlation coefficient of -0.34 found comparing Balanced Accuracy and Precision results across states) suggests avoiding relying on a single metric to evaluate models, and more specifically in this use case, to privilege Balanced Accuracy when the two strongly disagree because it provides a more comprehensive assessment of an algorithm's overall predictive ability.

\subsubsection{State-level Predictive Explanations}

\begin{figure}[!hbt]
    \centering
    \includegraphics[scale=0.4]{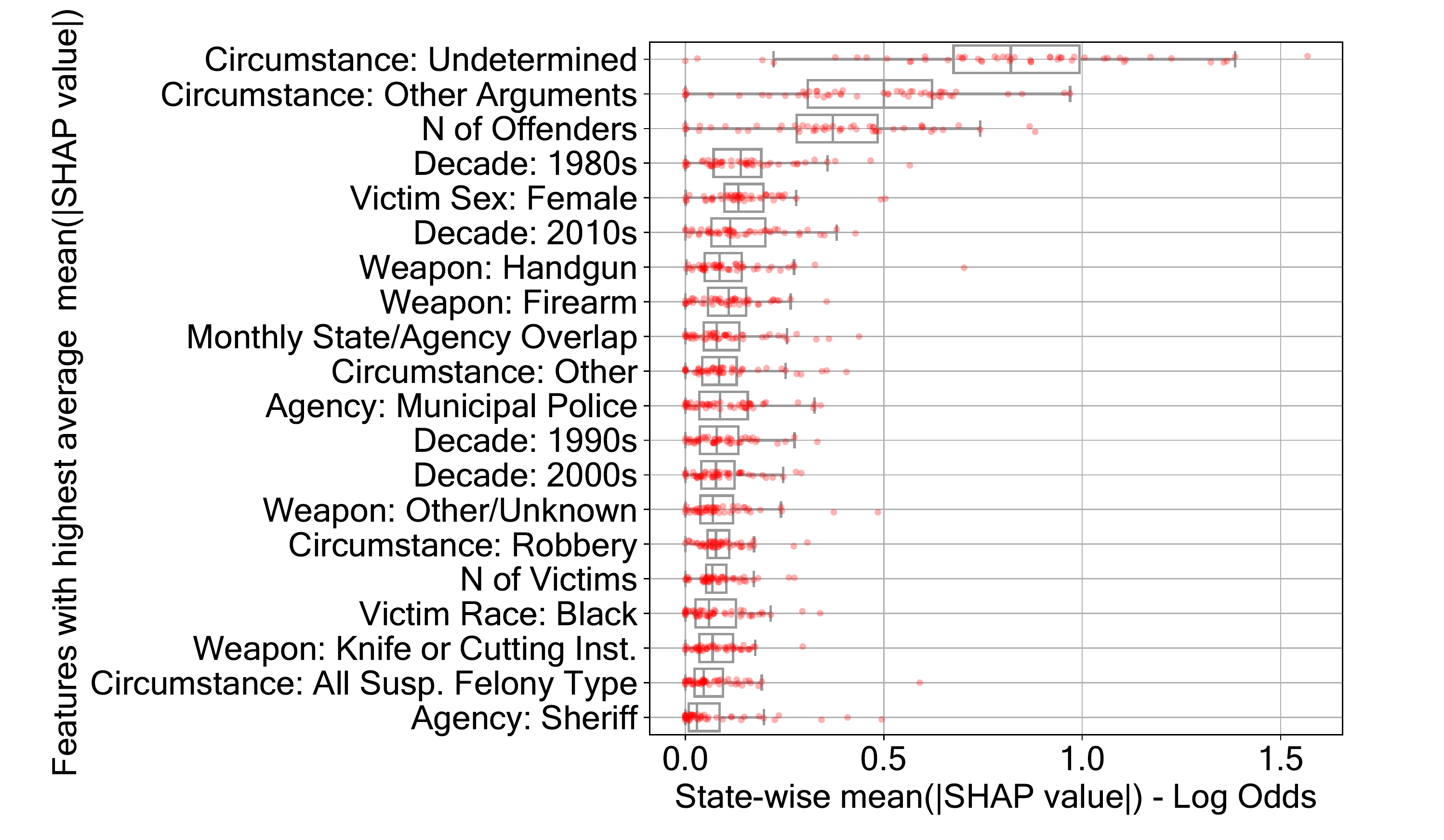}
    \caption{State-wise distribution of mean |SHAP value| in terms of log-odds for the twenty features with the highest average mean |SHAP value| across the 51 state datasets. The plot shows that the importance of these features vary consistently across states.}
    \label{fig:state_shap_pos}
\end{figure}

In the same vein as what was done for the national-level predictive model, state-level SHAP values have been computed using the best accurate XGBoost configuration for each state. In Figure \ref{fig:state_shap_pos} each point represents a state, and the plot displays the distribution of state-wise mean SHAP values in absolute terms, with the list of features indicating the twenty that obtain the highest average SHAP values in absolute terms after SHAP values have been derived state-wise. Although a sizable level of overlap can be appreciated, the list is different from the one seen in Figure \ref{absshap}, because estimates were gathered for each single state, rather than using all observation at once. Importantly, Figure \ref{fig:state_shap_pos} demonstrates how the overall magnitude of those features that are on average the most impactful across states varies significantly. This variance suggests that there is not a single universal and comprehensive model that can fit for all US states, further indicating how the same feature can have very different roles in explaining solved cases across jurisdictions. Feature "Circumstance: Undetermined", which was the most powerful at the national level, has a bandwidth of state-level magnitudes that goes from log-odds very close to zero, indicating almost no impact, to 1.5, indicating an extremely strong relationship with the outcome (note that the median is around 0.7, which was the mean SHAP value in absolute terms found in \ref{absshap}). Consistent variations, although with different distributions, occur for all features.

\begin{figure}[!hbt]
    \centering
    \includegraphics[scale=0.31]{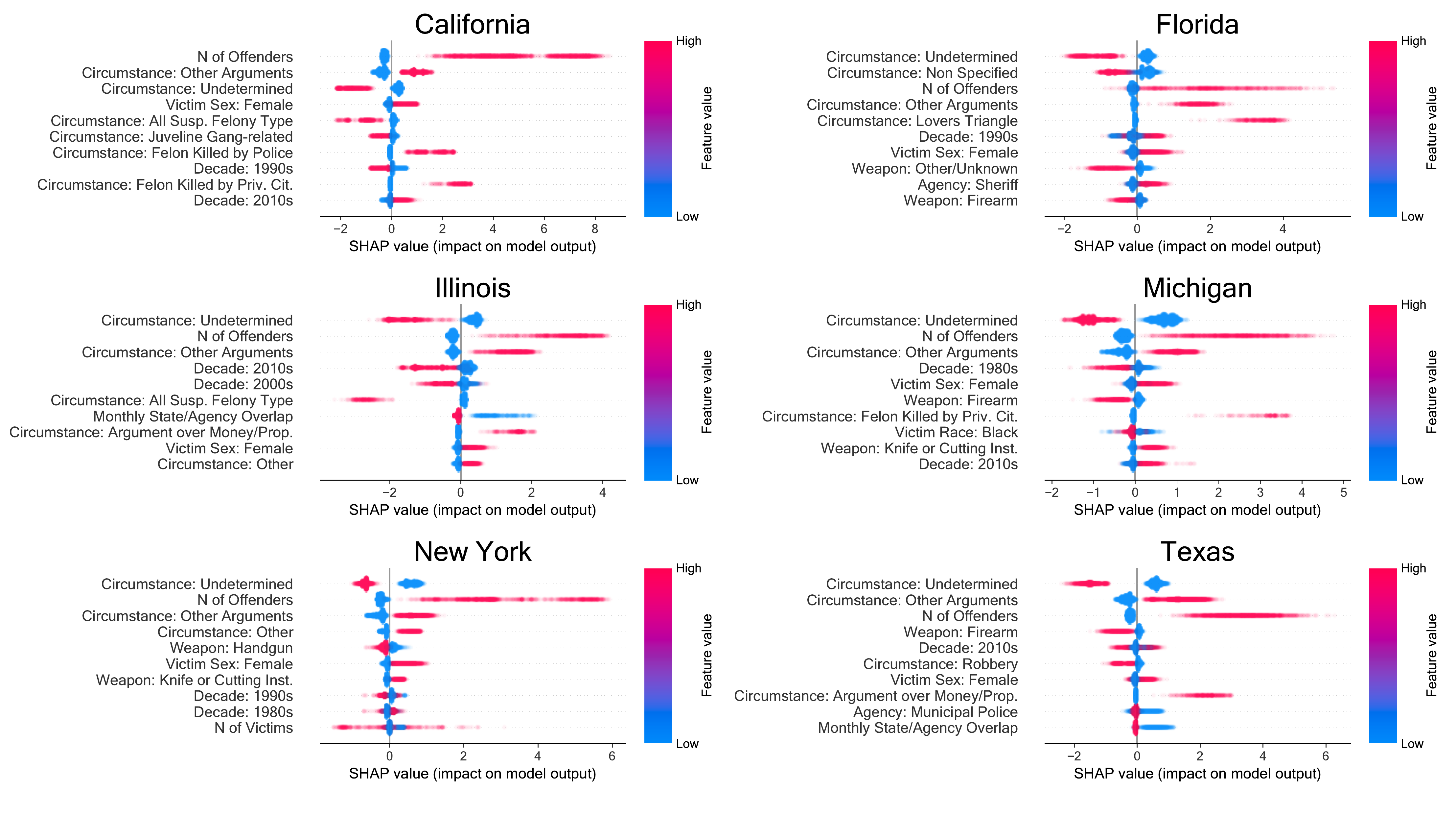}
    \caption{Distribution of SHAP values for the top 10 features at the state level for California (N=120,462), Florida (N=45,970), Illinois (N=35,298), Michigan (N=35,326), New York (N=59,516), and Texas (N=77,226), accounting for 46.44\% of the total homicides in the MAP dataset.}
    \label{fig:state_shap_neg}
\end{figure}

In light of the several mixed findings displayed in Figure \ref{fig:shap_dist}, and the further evidence of variation commented in Figure \ref{fig:state_shap_pos}, Figure \ref{fig:state_shap_neg} shows the distribution of raw SHAP values for six US states. The sample has been limited to the top 10 most impactful features for each of the six US states with the highest number of homicides in the MAP dataset, accounting for 46.44\% of total events: California (N=120,462), Florida (N=45,970), Illinois (N=35,298), Michigan (N=35,326), New York (N=59,516), and Texas (N=77,226). The first important thing to notice is that, compared to the national-level distribution seen in Figure \ref{fig:shap_dist}, the presence of mixed findings is less pronounced, supporting the hypothesis that national-level outcomes were hiding subnational patterns. 

The distribution of the features that demonstrated to be well-separated at the National level is also confirmed in the sample of state-level models here presented. "Circumstance: Undetermined" remains highly powerful in negatively predicting solved homicides. Similarly, "Circumstance: Other Arguments" has a strong positive role in predicting cleared cases. Homicides with female victims also keep their higher odds of being cleared. Additionally, Michigan, Florida, and Texas strengthen the finding that homicides perpetrated with firearms are correlated with unsolved cases (a similar outcome applies to New York, with homicides perpetrated with handguns leading to the same results). The same applies to the "Number of Offenders" feature.

Concerning those features that displayed mixed patterns at the national level, state-wise outcomes are generally much clearer in the direction of the magnitudes. One first example is homicides occurring in the context of robberies ("Circumstance: Robbery"), which in Texas (the only State that have the feature among the top 10 most impactful features) are associated with a reduction in the likelihood of a solved case.

Another interesting case is the distribution of SHAP values for juvenile gang-related homicides ("Circumstance: Juvenile Gang-related"), which displayed ambivalent directions at the national level, but in California are negatively associated with cleared cases, suggesting that California has struggled over the years, differently from other states, to investigate cases involving juveniles in gang-related criminal activities. 

Noteworthy, the mixed nature of findings for homicides with Black victims is much more straightforward in Michigan, where the odds of solving a case are substantially reduced if the victim is Black. Michigan is not the only state in which this pattern emerges. Figure \ref{fig:black_sup} in the Supplementary Materials shows a sample of other states (i.e., Connecticut, Iowa, Kansas, Massachusetts, Minnesota, and Missouri) where there is an evident negative relationship between Black victims and solved homicides.

\section{Discussion and Conclusions}
The present article is the first to explore the potential of machine learning for the study of homicide clearance and the promises of explainabilty methods for scrutinizing model predictions, addressing the long-standing problem of "black box" outcomes in computational methods. To conduct the analyses, the work has leveraged data from the Murder Accountability Project (MAP) on 804,751 homicides (corresponding to 768,431 separate events) that occurred from 1976 to 2019 in the United States \citep{HargroveMurderAccountabilityProject2019}. The MAP is a unique resource because it integrates events ordinarily recorded in the Supplementary Homicide Reports maintained by the FBI with more than 30,000 murders that were never officially registered into the UCR system and that were obtained through FOIA requests over the years. 

The work was framed in the debate concerning the underlying drivers of variations in homicide clearance, centered on the discretionary \citep{BlackBehaviorLaw1976, PaternosterProsecutorialDiscretionRequesting1984, PetersonChangingConceptionsRace1984} and nondiscretionary \citep{WolfgangPatternsCriminalHomicide1958, GottfredsonStudyBehaviorLaw1979,KlingerNegotiatingOrderPatrol1997a} theoretical perspectives that originally emerged in the Seventies and Fifties. Features included in the models encompassed as much as possible these two dimensions, combining information on offenders, victims, context and motives, and law enforcement. 

Before proceeding further, it should be noted however that this work comes with notable limitations.

First, the pool of variables available in the MAP dataset are not exhaustive of all the possible underlying correlates of homicide clearance. For instance, many more agency-level information would be required to better assess the relationship of organizational practices and resources with homicide clearance. Another missing variable is the time to clearance, which adds a further layer of information for studying solved cases, and has been previously addressed in the literature \citep{RegoecziClearingMurdersIt2008}. Additionally, the analysis completely lacks data on macrosocial dimensions that were part of Black's theory of law \citep{BlackBehaviorLaw1976}, including information on segregation, stratification, inequality, social control. The absence of such information does not allow us to have a fully comprehensive picture of the potential ecological, societal, economic and political intervening factors in explaining the observed variations, especially in such a long time frame. 

Second, while MAP does an excellent job in reducing the gap between registered homicides in the UCR system and those recorded by CDC, the present sample does not completely eliminate such difference. Cases that escaped the MAP dataset may be important for deriving predictive and explainability patterns, especially at the state level.

Third, MAP guidelines highlight that they estimate that 5-10\% out of total uncleared homicides are solved after data are collected in the SHR, and that such information is rarely updated accordingly in the SHR. Although such percentage is narrow, this issue warrants caution as some of the results, and those concerning predictive explainability foremost, may undergo variations. Nonetheless, the fact that most results here presented confirm patterns that were already found in other studies using data gathered from city- or county-level partnerships with police agencies, suggests that such issue has limited impact on the general trends reported in the paper. Furthermore, for both national and state models I have restricted the analysis of correlates to those features that were more robust (i.e., those that were ranked higher), assuming that minor non-random variations in the distribution of the target variable should not significantly alter their impact on homicide clearance. In addition, in the Supplementary Materials I have checked the robustness of the outcomes here presented by matching MAP data with a dataset on uncleared homicides compiled by \cite{WashingtonPostMurderImpunity2018}, and the majority of the most powerful drivers of clearance have been confirmed to hold also in such samples.

Fourth, the research design here presented does not allow to elaborate causal narratives for the obtained findings. All reported and commented results should be read in terms of correlation between a feature and the outcome of interest. This distinction is crucial, and must call for future endeavors aimed to study homicide clearance from a causality standpoint which still lacks in the quantitative scholarship on the topic. 

\color{black} Fifth, given the heterogeneity found at the state-level, it is reasonable to expect that further patterns could emerge from county-level and city-level analyses. This aspect calls for future work aimed at investigating more micro-level dynamics to verify the extent to which state themselves hide heterogeneity in terms of correlates of homicide clearance. \color{black}

Despite these limitations, this work offered a set of contributions, both methodologically and theoretically.

The present article sought to demonstrate the potential of machine learning for advancing our theoretical and practical knowledge on the long-standing issue of unsolved cases that exacerbate the widespread problem of homicide in the United States. Specifically, after evaluating nine algorithms across a total of 162 configurations through hyperparameter grid search at the national level, XGBoost \citep{ChenXGBoostScalableTree2016} - in its best configuration - has achieved the highest performance, slightly improving on Gradient Boosting Machines and Penalized Regression (i.e., Ridge, LASSO and Elastic Net), reaching cross-validated training Balanced Accuracy equal to $0.766\pm0.001$ and Precision equal to $0.863\pm0.001$. 

XGBoost has also been used as the algorithm to classify unsolved homicides at the state level, performing restricted grid-search to optimize accuracy in each dataset, fitting a total of 612 candidate configurations. Predictive performance varied consistently across states, with Balanced Accuracy ranging from 0.492 to 0.837, and Precision from 0.804 to 0.951. On the one hand, this degree of variance suggests that there may be heterogeneous patterns at play in each state jurisdiction. On the other hand, although useful and promising, machine learning should not be seen as a silver lining, calling for proper evaluation in each specific application context, as some mediocre results (i.e., North Dakota) highlight. 

Concerning algorithmic explainability, SHAP \citep{LundbergUnifiedApproachInterpreting2017a} was helpful in detecting predictors of homicide clearance, contributing to the extant literature. Notably, support for both the discretionary and nondiscretionary perspectives was found.

Several circumstances demonstrated to be highly relevant in explaining the probability that a homicide is cleared, in line with nondiscretionary prepositions. Undetermined circumstances are linked to lower odds of clearance, and the same applies to homicides that occurred in the context of concomitant felonies \citep{Leevaluelifedeath2005, AlderdenPredictingHomicideClearances2007a, LitwinDynamicNatureHomicide2007}. On the contrary, homicides perpetrated for reasons like personal revenge are easier to solve, similar to homicides that occurred for arguments over property and money and lovers-related issues. Other circumstances led instead to mixed findings at the national level, including those occurring in the context of juvenile-gang related activities and robberies. 

In line with the nondiscretionary frame, weapons also confirmed to be relevant predictors for clearance. Corroborating extant evidence \citep{PuckettFactorsAffectingHomicide2003, AddingtonUsingNationalIncidentBased2006, RobertsPredictorsHomicideClearance2007}, in fact, homicides committed with knives or cutting instruments are more likely to be solved, while the contrary holds for firearms and gunshots.

However, some findings also gave partial support to the discretionary perspective, in light of the importance of certain victim-level features in models' output. Clear patterns emerge indicating that homicides with female victims are more easily solved, in line with \cite{Leevaluelifedeath2005, AlderdenPredictingHomicideClearances2007a, RegoecziClearingMurdersIt2008}. The same holds for homicides targeting young victims, and particularly children aged 0-5, as already highlighted by previous works \citep{LitwinMultilevelMultivariateAnalysis2004, LitwinDynamicNatureHomicide2007}. Heterogeneous patterns at the national level instead emerge for the role of race and the number of victims in the same event. Detailed analyses on a sample of states, however, revealed the negative relationship between Black victims and cleared homicides. This is not only in line with part of the discretionary perspective, but also adds to the broader literature which repeatedly demonstrated racial disparities in the pipeline that goes from policing to the criminal justice system in the United States \citep{SampsonRacialEthnicDisparities1997, RehaviRacialDisparityFederal2014, Piersonlargescaleanalysisracial2020a, KnoxAdministrativeRecordsMask2020, FogliatoValidityArrestProxy2021a, DuxburyFearLoathingUnited2021}.

Regarding offender-related information, the number of perpetrators also influences predictive results, with more perpetrators raising the odds of clearance, somehow in connection with results from a previous study that detected shorter careers for multiple homicide offenders acting in teams \citep{CampedelliSurvivalRecidivisticRevealing2021}.

Finally, concerning the agency-level dimension, the concomitance of another homicide investigation by the same agency in the same jurisdiction in the same month decreases the likelihood of solved homicides, with some residual exceptions. This finding might reflect excessive burden and workload, especially for small agencies that do not to investigate homicides frequently. Furthermore, homicides investigated by municipal police agencies seem to have lower clearance odds than alternative agencies.

In general, results offer both clear and mixed outcomes. This mix of positive and negative SHAP values at the national level characterizing many important features (e.g., juvenile gang-related circumstances, robbery circumstances, race of the victim) may be due to heterogeneous sub-patterns in the data.

State-level SHAP models, in fact, support the hypothesis that national-scale models hide more localized dynamics given that the analysis reveals substantial variation in overall feature importance across state jurisdictions. Particularly, a specific analysis of the six states with the highest number of homicides (i.e., California, Florida, Illinois, Michigan, New York, and Texas) demonstrated that those which were mixed findings at the national level are much clearer when considering a meso-level analytical perspective. 

While, for instance, juvenile gang-related homicides had heterogeneous results at the national level, in California homicides with such characteristics are only related to lower odds of clearance. 

Importantly, the role of race also appeared less blurred at the state level. With few exceptions, for instance, homicides with Black victims were associated with lower odds of clearance in Michigan, in line with some previous studies \citep{Leevaluelifedeath2005, RegoecziClearingMurdersIt2008, AlderdenPredictingHomicideClearances2007a} and in support to the discretionary perspective. A supplementary analysis also revealed the same finding for six states, namely Connecticut, Iowa, Kansas, Massachusetts, Minnesota, and Missouri. 

Overall, the nature of the findings and the contribution of state-level analyses indicate that relying on a single theoretical perspective is rather limiting. The complex patterns that emerge, their temporal and geographical variations, and the simultaneous support for discretionary and nondiscretionary proxies call for theoretical revisiting efforts.

From a practical point of view, state-level variability prompts ad-hoc focused interventions and policies to improve performance for subtypes of homicides, e.g., those involving juvenile gang-related violence in California, or those involving Black victims in Michigan, Connecticut, Iowa, Kansas, Massachusetts, Minnesota and Missouri. No one-size-fits-all strategy can be enginereed for all US states, and tailored interventions to reduce the rate of uncleared cases are hence imperative.

Concerning future research, as anticipated, scholarly efofrts should emphasize the importance of causal research designs to offer even stronger evidence for performance improvement. \color{black} Future endeavors must also enrich the information portfolio available in the MAP dataset with organizational data on police activity and macro-level social factors.   This evolution should go in parallel with efforts to ameliorate reporting practices and information completeness. Particularly, the shift towards the NIBRS system will represent a significant opportunity for police departments, analysts and crime researchers in the decades to come, given the wider set of information associated with each registered offense  --- including homicides. In the future, data gathering practices for major crimes as homicides could also benefit from the inclusion of additional variables, such as those reporting forensic information, which may be important predictors of investigation outcomes, as well as from easier integration with other agency-level information. \color{black}

\section{Declaration of Interest}
None.

\section{Acknowledgments}
I thank Thomas Hargrove, the founder of the Murder Accountability Project, for providing me with precious detail regarding the data used to carry out this work. I also thank three anonymous referees for their comments and recommendations on an earlier version of this manuscript.

\section{Data and Code availability}
All datasets used in this study are available from the Murder Accountability Project Website (\href{http://www.murderdata.org/}{http://www.murderdata.org/}). Code is publicly available at the following \href{https://github.com/gcampede/Interpretable-ML-for-Homicide-Clearance}{GitHub} repository.

\color{black}
\newpage

%\printbibliography
\bibliography{ML_Homicides}

\begin{thebibliography}{}

\bibitem[Addington, 2006]{AddingtonUsingNationalIncidentBased2006}
Addington, L.~A. (2006).
\newblock Using {National} {Incident}-{Based} {Reporting} {System} {Murder}
  {Data} to {Evaluate} {Clearance} {Predictors}: {A} {Research} {Note}.
\newblock {\em Homicide Studies}, 10(2):140--152.
\newblock Publisher: SAGE Publications Inc.

\bibitem[Addington, 2007]{AddingtonHotvsCold2007}
Addington, L.~A. (2007).
\newblock Hot vs. {Cold} {Cases}: {Examining} {Time} to {Clearance} for
  {Homicides} {Using} {NIBRS} {Data}.
\newblock {\em Justice Research and Policy}, 9(2):87--112.
\newblock Publisher: SAGE Publications Inc.

\bibitem[Alderden and Lavery, 2007]{AlderdenPredictingHomicideClearances2007a}
Alderden, M.~A. and Lavery, T.~A. (2007).
\newblock Predicting {Homicide} {Clearances} in {Chicago}: {Investigating}
  {Disparities} in {Predictors} {Across} {Different} {Types} of {Homicide}.
\newblock {\em Homicide Studies}, 11(2):115--132.
\newblock Publisher: SAGE Publications Inc.

\bibitem[Angelov et~al., 2021]{AngelovExplainableartificialintelligence2021}
Angelov, P.~P., Soares, E.~A., Jiang, R., Arnold, N.~I., and Atkinson, P.~M.
  (2021).
\newblock Explainable artificial intelligence: an analytical review.
\newblock {\em WIREs Data Mining and Knowledge Discovery}, 11(5):e1424.
\newblock \_eprint: https://onlinelibrary.wiley.com/doi/pdf/10.1002/widm.1424.

\bibitem[Armour, 2002]{ArmourExperiencesCovictimsHomicide2002}
Armour, M.~P. (2002).
\newblock Experiences of {Covictims} of {Homicide}: {Implications} for
  {Research} and {Practice}.
\newblock {\em Trauma, Violence, \& Abuse}, 3(2):109--124.
\newblock Publisher: SAGE Publications.

\bibitem[Asher, 2021]{AsherMurderRoseAlmost2021}
Asher, J. (2021).
\newblock Murder {Rose} by {Almost} 30\% in 2020. {It}’s {Rising} at a
  {Slower} {Rate} in 2021.
\newblock {\em The New York Times}.

\bibitem[Barber et~al., 2002]{BarberUnderestimatesunintentionalfirearm2002}
Barber, C., Hemenway, D., Hochstadt, J., and Azrael, D. (2002).
\newblock Underestimates of unintentional firearm fatalities: comparing
  {Supplementary} {Homicide} {Report} data with the {National} {Vital}
  {Statistics} {System}.
\newblock {\em Injury Prevention}, 8(3):252--256.
\newblock Publisher: BMJ Publishing Group Ltd Section: METHODOLOGIC ISSUES.

\bibitem[Baskin and Sommers, 2010]{Baskininfluenceforensicevidence2010}
Baskin, D. and Sommers, I. (2010).
\newblock The influence of forensic evidence on the case outcomes of homicide
  incidents.
\newblock {\em Journal of Criminal Justice}, 38(6):1141--1149.

\bibitem[Berk, 2013]{BerkAlgorithmiccriminology2013d}
Berk, R. (2013).
\newblock Algorithmic criminology.
\newblock {\em Security Informatics}, 2(1):5.

\bibitem[Black, 1976]{BlackBehaviorLaw1976}
Black, D. (1976).
\newblock {\em The {Behavior} of {Law}}.
\newblock Academic Press, New York, NY.

\bibitem[Boser et~al., 1992]{Bosertrainingalgorithmoptimal1992}
Boser, B.~E., Guyon, I.~M., and Vapnik, V.~N. (1992).
\newblock A training algorithm for optimal margin classifiers.
\newblock In {\em Proceedings of the fifth annual workshop on {Computational}
  learning theory}, {COLT} '92, pages 144--152, New York, NY, USA. Association
  for Computing Machinery.

\bibitem[Braga, 2021]{BragaImprovingPoliceClearance2021}
Braga, A.~A. (2021).
\newblock Improving {Police} {Clearance} {Rates} of {Shootings}: {A} {Review}
  of the {Evidence}.
\newblock Technical report, Manhattan Institute.

\bibitem[Braga and Dusseault, 2018]{BragaCanHomicideDetectives2018}
Braga, A.~A. and Dusseault, D. (2018).
\newblock Can {Homicide} {Detectives} {Improve} {Homicide} {Clearance} {Rates}?
\newblock {\em Crime \& Delinquency}, 64(3):283--315.
\newblock Publisher: SAGE Publications Inc.

\bibitem[Braga et~al., 1999]{BragaYouthHomicideBoston1999}
Braga, A.~A., Piehl, A.~M., and Kennedy, D.~M. (1999).
\newblock Youth {Homicide} in {Boston}: {An} {Assessment} of {Supplementary}
  {Homicide} {Report} {Data}.
\newblock {\em Homicide Studies}, 3(4):277--299.
\newblock Publisher: SAGE Publications Inc.

\bibitem[Breiman, 2001]{BreimanRandomForests2001c}
Breiman, L. (2001).
\newblock Random {Forests}.
\newblock {\em Machine Learning}, 45(1):5--32.

\bibitem[Brennan and Oliver, 2013]{BrennanEmergenceMachineLearning2013c}
Brennan, T. and Oliver, W.~L. (2013).
\newblock The {Emergence} of {Machine} {Learning} {Techniques} in
  {Criminology}: {Implications} of {Complexity} in our {Data} and in {Research}
  {Questions}.
\newblock {\em Criminology \& Public Policy}, 12(3):551--562.

\bibitem[Browne and Williams, 1993]{BrowneGENDERINTIMACYLETHAL1993}
Browne, A. and Williams, K.~R. (1993).
\newblock Gender, {Intimacy} and {Lethal} {Violence}: {Trends} from 1976
  through 1987.
\newblock {\em Gender \& Society}, 7(1):78--98.

\bibitem[Buitinck et~al., 2013]{BuitinckAPIdesignmachine2013}
Buitinck, L., Louppe, G., Blondel, M., Pedregosa, F., Mueller, A., Grisel, O.,
  Niculae, V., Prettenhofer, P., Gramfort, A., Grobler, J., Layton, R.,
  Vanderplas, J., Joly, A., Holt, B., and Varoquaux, G. (2013).
\newblock {API} design for machine learning software: experiences from the
  scikit-learn project.
\newblock {\em arXiv:1309.0238 [cs]}.
\newblock arXiv: 1309.0238.

\bibitem[Campedelli, 2021]{CampedelliWherearewe2021}
Campedelli, G.~M. (2021).
\newblock Where are we? {Using} {Scopus} to map the literature at the
  intersection between artificial intelligence and research on crime.
\newblock {\em Journal of Computational Social Science}, 4(2):503--530.

\bibitem[Campedelli and Yaksic,
  2021]{CampedelliSurvivalRecidivisticRevealing2021}
Campedelli, G.~M. and Yaksic, E. (2021).
\newblock Survival of the {Recidivistic}? {Revealing} {Factors} {Associated}
  with the {Criminal} {Career} {Length} of {Multiple} {Homicide} {Offenders}.
\newblock {\em Homicide Studies}, page 10887679211010882.
\newblock Publisher: SAGE Publications Inc.

\bibitem[Cardarelli and Cavanagh, 1994]{CardarelliUnclearedHomicidesUnited1994}
Cardarelli, A. and Cavanagh, D. (1994).
\newblock Uncleared {Homicides} in the {United} {States}: {AnExploratory}
  {Study} of {Trends} and {Patterns}.
\newblock Technical report.

\bibitem[Carter and Carter, 2016]{CarterEffectivePoliceHomicide2016}
Carter, D.~L. and Carter, J.~G. (2016).
\newblock Effective {Police} {Homicide} {Investigations}: {Evidence} {From}
  {Seven} {Cities} {With} {High} {Clearance} {Rates}.
\newblock {\em Homicide Studies}, 20(2):150--176.
\newblock Publisher: SAGE Publications Inc.

\bibitem[Chalfin et~al., 2020]{ChalfinPoliceForceSize2020}
Chalfin, A., Hansen, B., Weisburst, E.~K., and Williams, Morgan~C., J. (2020).
\newblock Police {Force} {Size} and {Civilian} {Race}.
\newblock Working {Paper} 28202, National Bureau of Economic Research.
\newblock Series: Working Paper Series.

\bibitem[Chan and Beauregard, 2016]{ChanChoiceWeaponWeapon2016}
Chan, H. C.~O. and Beauregard, E. (2016).
\newblock Choice of {Weapon} or {Weapon} of {Choice}? {Examining} the
  {Interactions} between {Victim} {Characteristics} in {Single}-victim {Male}
  {Sexual} {Homicide} {Offenders}.
\newblock {\em Journal of Investigative Psychology and Offender Profiling},
  13(1):70--88.
\newblock \_eprint: https://onlinelibrary.wiley.com/doi/pdf/10.1002/jip.1432.

\bibitem[Chen and Guestrin, 2016]{ChenXGBoostScalableTree2016}
Chen, T. and Guestrin, C. (2016).
\newblock {XGBoost}: {A} {Scalable} {Tree} {Boosting} {System}.
\newblock {\em Proceedings of the 22nd ACM SIGKDD International Conference on
  Knowledge Discovery and Data Mining}, pages 785--794.
\newblock arXiv: 1603.02754.

\bibitem[{Council on Criminal Justice},
  2021]{CouncilonCriminalJusticeHomicideTrendsWhat2021}
{Council on Criminal Justice} (2021).
\newblock Homicide {Trends}: {What} {You} {Need} to {Know}.

\bibitem[Datta et~al., 2016]{DattaAlgorithmicTransparencyQuantitative2016}
Datta, A., Sen, S., and Zick, Y. (2016).
\newblock Algorithmic {Transparency} via {Quantitative} {Input} {Influence}:
  {Theory} and {Experiments} with {Learning} {Systems}.
\newblock In {\em 2016 {IEEE} {Symposium} on {Security} and {Privacy} ({SP})},
  pages 598--617.
\newblock ISSN: 2375-1207.

\bibitem[DeLisi et~al., 2010]{DeLisiMurdernumbersmonetary2010}
DeLisi, M., Kosloski, A., Sween, M., Hachmeister, E., Moore, M., and Drury, A.
  (2010).
\newblock Murder by numbers: monetary costs imposed by a sample of homicide
  offenders.
\newblock {\em The Journal of Forensic Psychiatry \& Psychology},
  21(4):501--513.
\newblock Publisher: Routledge \_eprint:
  https://doi.org/10.1080/14789940903564388.

\bibitem[DeLisi et~al., 2016]{DeLisiUnpredictabilityMurderJuvenile2016}
DeLisi, M., Piquero, A.~R., and Cardwell, S.~M. (2016).
\newblock The {Unpredictability} of {Murder}: {Juvenile} {Homicide} in the
  {Pathways} to {Desistance} {Study}.
\newblock {\em Youth Violence and Juvenile Justice}, 14(1):26--42.
\newblock Publisher: SAGE Publications.

\bibitem[Duxbury, 2021]{DuxburyFearLoathingUnited2021}
Duxbury, S.~W. (2021).
\newblock Fear or {Loathing} in the {United} {States}? {Public} {Opinion} and
  the {Rise} of {Racial} {Disparity} in {Mass} {Incarceration}, 1978–2015.
\newblock {\em Social Forces}, 100(2):427--453.

\bibitem[Elgar and Aitken, 2011]{ElgarIncomeinequalitytrust2011}
Elgar, F.~J. and Aitken, N. (2011).
\newblock Income inequality, trust and homicide in 33 countries.
\newblock {\em European Journal of Public Health}, 21(2):241--246.

\bibitem[Fogliato et~al., 2021]{FogliatoValidityArrestProxy2021a}
Fogliato, R., Xiang, A., Lipton, Z., Nagin, D., and Chouldechova, A. (2021).
\newblock On the {Validity} of {Arrest} as a {Proxy} for {Offense}: {Race} and
  the {Likelihood} of {Arrest} for {Violent} {Crimes}.
\newblock {\em arXiv:2105.04953 [stat]}.
\newblock arXiv: 2105.04953.

\bibitem[Fox and Levin, 1991]{FoxHomicideElderlyResearch1991}
Fox, J.~A. and Levin, J. (1991).
\newblock Homicide {Against} the {Elderly}: {A} {Research} {Note}*.
\newblock {\em Criminology}, 29(2):317--327.
\newblock \_eprint:
  https://onlinelibrary.wiley.com/doi/pdf/10.1111/j.1745-9125.1991.tb01069.x.

\bibitem[Friedman, 2001]{FriedmanGreedyfunctionapproximation2001}
Friedman, J.~H. (2001).
\newblock Greedy function approximation: {A} gradient boosting machine.
\newblock {\em The Annals of Statistics}, 29(5).

\bibitem[Gallup-Black, 2005]{Gallup-BlackTwentyYearsRural2005}
Gallup-Black, A. (2005).
\newblock Twenty {Years} of {Rural} and {Urban} {Trends} in {Family} and
  {Intimate} {Partner} {Homicide}: {Does} {Place} {Matter}?
\newblock {\em Homicide Studies}, 9(2):149--173.
\newblock Publisher: SAGE Publications Inc.

\bibitem[Gottfredson and Hindelang, 1979]{GottfredsonStudyBehaviorLaw1979}
Gottfredson, M.~R. and Hindelang, M.~J. (1979).
\newblock A {Study} of the {Behavior} of {Law}.
\newblock {\em American Sociological Review}, 44(1):3--18.
\newblock Publisher: [American Sociological Association, Sage Publications,
  Inc.].

\bibitem[Gunning et~al., 2019]{GunningXAIExplainableartificial2019}
Gunning, D., Stefik, M., Choi, J., Miller, T., Stumpf, S., and Yang, G.-Z.
  (2019).
\newblock {XAI}—{Explainable} artificial intelligence.
\newblock {\em Science Robotics}.
\newblock Publisher: American Association for the Advancement of Science.

\bibitem[Hargrove, 2019]{HargroveMurderAccountabilityProject2019}
Hargrove, T. (2019).
\newblock Murder {Accountability} {Project}.

\bibitem[Hastie et~al., 2009]{HastieElementsStatisticalLearning2009}
Hastie, T., Tibshirani, R., and Friedman, J. (2009).
\newblock {\em The {Elements} {Of} {Statistical} {Learning}: {Data} {Mining},
  {Inference}, {And} {Prediction}, {Second} {Edition}}.
\newblock Springer Nature, New York, NY, 2° edizione edition.

\bibitem[Holzinger et~al., 2017]{HolzingerWhatweneed2017}
Holzinger, A., Biemann, C., Pattichis, C.~S., and Kell, D.~B. (2017).
\newblock What do we need to build explainable {AI} systems for the medical
  domain?
\newblock {\em arXiv:1712.09923 [cs, stat]}.
\newblock arXiv: 1712.09923.

\bibitem[Kaplan, 2020]{KaplanChapterSupplementaryHomicide2020}
Kaplan, J. (2020).
\newblock {\em Chapter 6 {Supplementary} {Homicide} {Reports} ({SHR})
  {\textbar} {Uniform} {Crime} {Reporting} ({UCR}) {Program} {Data}: {A}
  {Practitioner}’s {Guide}}.

\bibitem[Keel et~al., 2009]{KeelExploratoryAnalysisFactors2009}
Keel, T.~G., Jarvis, J.~P., and Muirhead, Y.~E. (2009).
\newblock An {Exploratory} {Analysis} of {Factors} {Affecting} {Homicide}
  {Investigations}: {Examining} the {Dynamics} of {Murder} {Clearance} {Rates}.
\newblock {\em Homicide Studies}, 13(1):50--68.
\newblock Publisher: SAGE Publications Inc.

\bibitem[Kennedy et~al., 2021]{KennedyEnvironmentalFactorsInfluencing2021}
Kennedy, L.~W., Caplan, J.~M., Piza, E.~L., and Thomas, A.~L. (2021).
\newblock Environmental {Factors} {Influencing} {Urban} {Homicide} {Clearance}
  {Rates}: {A} {Spatial} {Analysis} of {New} {York} {City}.
\newblock {\em Homicide Studies}, 25(4):313--334.
\newblock Publisher: SAGE Publications Inc.

\bibitem[Klinger, 1997]{KlingerNegotiatingOrderPatrol1997a}
Klinger, D.~A. (1997).
\newblock Negotiating {Order} in {Patrol} {Work}: {An} {Ecological} {Theory} of
  {Police} {Response} to {Deviance}*.
\newblock {\em Criminology}, 35(2):277--306.
\newblock \_eprint:
  https://onlinelibrary.wiley.com/doi/pdf/10.1111/j.1745-9125.1997.tb00877.x.

\bibitem[Knox et~al., 2020]{KnoxAdministrativeRecordsMask2020}
Knox, D., Lowe, W., and Mummolo, J. (2020).
\newblock Administrative {Records} {Mask} {Racially} {Biased} {Policing}.
\newblock {\em American Political Science Review}, 114(3):619--637.
\newblock Publisher: Cambridge University Press.

\bibitem[Lee, 2005]{Leevaluelifedeath2005}
Lee, C. (2005).
\newblock The value of life in death: {Multiple} regression and event history
  analyses of homicide clearance in {Los} {Angeles} {County}.
\newblock {\em Journal of Criminal Justice}, 33(6):527--534.

\bibitem[Litwin, 2004]{LitwinMultilevelMultivariateAnalysis2004}
Litwin, K.~J. (2004).
\newblock A {Multilevel} {Multivariate} {Analysis} of {Factors} {Affecting}
  {Homicide} {Clearances}.
\newblock {\em Journal of Research in Crime and Delinquency}, 41(4):327--351.
\newblock Publisher: SAGE Publications Inc.

\bibitem[Litwin and Xu, 2007]{LitwinDynamicNatureHomicide2007}
Litwin, K.~J. and Xu, Y. (2007).
\newblock The {Dynamic} {Nature} of {Homicide} {Clearances}: {A} {Multilevel}
  {Model} {Comparison} of {Three} {Time} {Periods}.
\newblock {\em Homicide Studies}, 11(2):94--114.
\newblock Publisher: SAGE Publications Inc.

\bibitem[Loftin et~al., 2015]{LoftinAccuracySupplementaryHomicide2015}
Loftin, C., McDowall, D., Curtis, K., and Fetzer, M.~D. (2015).
\newblock The {Accuracy} of {Supplementary} {Homicide} {Report} {Rates} for
  {Large} {U}.{S}. {Cities}.
\newblock {\em Homicide Studies}, 19(1):6--27.
\newblock Publisher: SAGE Publications Inc.

\bibitem[Lundberg and Lee, 2017a]{LundbergUnifiedApproachInterpreting2017}
Lundberg, S. and Lee, S.-I. (2017a).
\newblock A {Unified} {Approach} to {Interpreting} {Model} {Predictions}.
\newblock {\em arXiv:1705.07874 [cs, stat]}.
\newblock arXiv: 1705.07874.

\bibitem[Lundberg and Lee, 2017b]{LundbergUnifiedApproachInterpreting2017a}
Lundberg, S. and Lee, S.-I. (2017b).
\newblock A {Unified} {Approach} to {Interpreting} {Model} {Predictions}.
\newblock {\em arXiv:1705.07874 [cs, stat]}.
\newblock arXiv: 1705.07874.

\bibitem[Lundberg et~al., 2019]{LundbergExplainableAITrees2019}
Lundberg, S.~M., Erion, G., Chen, H., DeGrave, A., Prutkin, J.~M., Nair, B.,
  Katz, R., Himmelfarb, J., Bansal, N., and Lee, S.-I. (2019).
\newblock Explainable {AI} for {Trees}: {From} {Local} {Explanations} to
  {Global} {Understanding}.
\newblock {\em arXiv:1905.04610 [cs, stat]}.
\newblock arXiv: 1905.04610.

\bibitem[Lundman and Myers, 2012]{LundmanExplanationsHomicideClearances2012}
Lundman, R.~J. and Myers, M. (2012).
\newblock Explanations of {Homicide} {Clearances}: {Do} {Results} {Vary}
  {Dependent} {Upon} {Operationalization} and {Initial} ({Time} 1) and
  {Updated} ({Time} 2) {Data}?
\newblock {\em Homicide Studies}, 16(1):23--40.
\newblock Publisher: SAGE Publications Inc.

\bibitem[Mancik and Parker, 2019]{MancikHomicideclearancespre2019}
Mancik, A.~M. and Parker, K.~F. (2019).
\newblock Homicide clearances during pre- and post-{U}.{S}. crime drop eras:
  the role of structural predictors and demographic shifts, 1976–2015.
\newblock {\em Journal of Crime and Justice}, 42(3):237--256.
\newblock Publisher: Routledge \_eprint:
  https://doi.org/10.1080/0735648X.2018.1526101.

\bibitem[Mancik et~al., 2018]{MancikNeighborhoodContextHomicide2018a}
Mancik, A.~M., Parker, K.~F., and Williams, K.~R. (2018).
\newblock Neighborhood {Context} and {Homicide} {Clearance}: {Estimating} the
  {Effects} of {Collective} {Efficacy}.
\newblock {\em Homicide Studies}, 22(2):188--213.
\newblock Publisher: SAGE Publications Inc.

\bibitem[Mastrocinque et~al., 2015]{MastrocinqueStillLeftHere2015}
Mastrocinque, J.~M., Metzger, J.~W., Madeira, J., Lang, K., Pruss, H.,
  Navratil, P.~K., Sandys, M., and Cerulli, C. (2015).
\newblock I’m {Still} {Left} {Here} {With} the {Pain}: {Exploring} the
  {Health} {Consequences} of {Homicide} on {Families} and {Friends}.
\newblock {\em Homicide Studies}, 19(4):326--349.

\bibitem[McDowall and Curtis, 2015]{McDowallSeasonalVariationHomicide2015}
McDowall, D. and Curtis, K.~M. (2015).
\newblock Seasonal {Variation} in {Homicide} and {Assault} {Across} {Large}
  {U}.{S}. {Cities}.
\newblock {\em Homicide Studies}, 19(4):303--325.
\newblock Publisher: SAGE Publications Inc.

\bibitem[McEwen and Regoeczi, 2015]{McEwenForensicEvidenceHomicide2015}
McEwen, T. and Regoeczi, W. (2015).
\newblock Forensic {Evidence} in {Homicide} {Investigations} and
  {Prosecutions}.
\newblock {\em Journal of Forensic Sciences}, 60(5):1188--1198.

\bibitem[Ousey and Lee, 2010]{OuseyKnowUnknownDecline2010}
Ousey, G.~C. and Lee, M.~R. (2010).
\newblock To {Know} the {Unknown}: {The} {Decline} in {Homicide} {Clearance}
  {Rates}, 1980—2000.
\newblock {\em Criminal Justice Review}, 35(2):141--158.
\newblock Publisher: SAGE Publications Inc.

\bibitem[Papachristos and Wildeman,
  2014]{PapachristosNetworkExposureHomicide2014}
Papachristos, A.~V. and Wildeman, C. (2014).
\newblock Network {Exposure} and {Homicide} {Victimization} in an {African}
  {American} {Community}.
\newblock {\em American Journal of Public Health}, 104(1):143--150.
\newblock Publisher: American Public Health Association.

\bibitem[Paternoster, 1984]{PaternosterProsecutorialDiscretionRequesting1984}
Paternoster, R. (1984).
\newblock Prosecutorial {Discretion} in {Requesting} the {Death} {Penalty}: {A}
  {Case} of {Victim}-{Based} {Racial} {Discrimination}.
\newblock {\em Law \& Society Review}, 18(3):437--478.
\newblock Publisher: [Wiley, Law and Society Association].

\bibitem[Peterson and Hagan, 1984]{PetersonChangingConceptionsRace1984}
Peterson, R.~D. and Hagan, J. (1984).
\newblock Changing {Conceptions} of {Race}: {Towards} an {Account} of
  {Anomalous} {Findings} of {Sentencing} {Research}.
\newblock {\em American Sociological Review}, 49(1):56--70.
\newblock Publisher: [American Sociological Association, Sage Publications,
  Inc.].

\bibitem[Pierson et~al., 2020]{Piersonlargescaleanalysisracial2020a}
Pierson, E., Simoiu, C., Overgoor, J., Corbett-Davies, S., Jenson, D.,
  Shoemaker, A., Ramachandran, V., Barghouty, P., Phillips, C., Shroff, R., and
  Goel, S. (2020).
\newblock A large-scale analysis of racial disparities in police stops across
  the {United} {States}.
\newblock {\em Nature Human Behaviour}, 4(7):736--745.
\newblock Bandiera\_abtest: a Cg\_type: Nature Research Journals Number: 7
  Primary\_atype: Research Publisher: Nature Publishing Group Subject\_term:
  Computer science;Criminology;Sociology;Statistics Subject\_term\_id:
  computer-science;criminology;sociology;statistics.

\bibitem[Pizarro et~al., 2020]{PizarroImpactInvestigationStrategies2020}
Pizarro, J.~M., Terrill, W., and LoFaso, C.~A. (2020).
\newblock The {Impact} of {Investigation} {Strategies} and {Tactics} on
  {Homicide} {Clearance}.
\newblock {\em Homicide Studies}, 24(1):3--24.
\newblock Publisher: SAGE Publications Inc.

\bibitem[Puckett and Lundman, 2003]{PuckettFactorsAffectingHomicide2003}
Puckett, J.~L. and Lundman, R.~J. (2003).
\newblock Factors {Affecting} {Homicide} {Clearances}: {Multivariate}
  {Analysis} {Of} {A} {More} {Complete} {Conceptual} {Framework}.
\newblock {\em Journal of Research in Crime and Delinquency}, 40(2):171--193.
\newblock Publisher: SAGE Publications Inc.

\bibitem[Quinlan, 1986]{QuinlanInductiondecisiontrees1986}
Quinlan, J.~R. (1986).
\newblock Induction of decision trees.
\newblock {\em Machine Learning}, 1(1):81--106.

\bibitem[Redelings et~al., 2010]{RedelingsYearsYourLife2010}
Redelings, M., Lieb, L., and Sorvillo, F. (2010).
\newblock Years off {Your} {Life}? {The} {Effects} of {Homicide} on {Life}
  {Expectancy} by {Neighborhood} and {Race}/{Ethnicity} in {Los} {Angeles}
  {County}.
\newblock {\em Journal of Urban Health : Bulletin of the New York Academy of
  Medicine}, 87(4):670--676.

\bibitem[Regoeczi et~al., 2020]{RegoecziHomicideInvestigationsContext2020}
Regoeczi, W.~C., Jarvis, J., and Mancik, A. (2020).
\newblock Homicide {Investigations} in {Context}: {Exploring} {Explanations}
  for the {Divergent} {Impacts} of {Victim} {Race}, {Gender}, {Elderly}
  {Victims}, and {Firearms} on {Homicide} {Clearances}.
\newblock {\em Homicide Studies}, 24(1):25--44.
\newblock Publisher: SAGE Publications Inc.

\bibitem[Regoeczi et~al., 2008]{RegoecziClearingMurdersIt2008}
Regoeczi, W.~C., Jarvis, J., and Riedel, M. (2008).
\newblock Clearing {Murders}: {Is} {It} about {Time}?
\newblock {\em Journal of Research in Crime and Delinquency}, 45(2):142--162.
\newblock Publisher: SAGE Publications Inc.

\bibitem[Regoeczi et~al., 2000]{RegoecziUnclearedHomicidesCanada2000}
Regoeczi, W.~C., Kennedy, L.~W., and Silverman, R.~A. (2000).
\newblock Uncleared {Homicides}: {A} {Canada}/{United} {States} {Comparison}.
\newblock {\em Homicide Studies}, 4(2):135--161.
\newblock Publisher: SAGE Publications Inc.

\bibitem[Rehavi and Starr, 2014]{RehaviRacialDisparityFederal2014}
Rehavi, M.~M. and Starr, S.~B. (2014).
\newblock Racial {Disparity} in {Federal} {Criminal} {Sentences}.
\newblock {\em Journal of Political Economy}, 122(6):1320--1354.
\newblock Publisher: The University of Chicago Press.

\bibitem[Ribeiro et~al., 2016]{RibeiroWhyShouldTrust2016a}
Ribeiro, M.~T., Singh, S., and Guestrin, C. (2016).
\newblock "{Why} {Should} {I} {Trust} {You}?": {Explaining} the {Predictions}
  of {Any} {Classifier}.
\newblock {\em arXiv:1602.04938 [cs, stat]}.
\newblock arXiv: 1602.04938.

\bibitem[Riedel, 2008]{RiedelHomicideArrestClearances2008a}
Riedel, M. (2008).
\newblock Homicide {Arrest} {Clearances}: {A} {Review} of the {Literature}.
\newblock {\em Sociology Compass}, 2(4):1145--1164.
\newblock \_eprint:
  https://onlinelibrary.wiley.com/doi/pdf/10.1111/j.1751-9020.2008.00130.x.

\bibitem[Riedel and Jarvis, 1999]{RiedelDeclineArrestClearances1999}
Riedel, M. and Jarvis, J. (1999).
\newblock The {Decline} of {Arrest} {Clearances} for {Criminal} {Homicide}:
  {Causes}, {Correlates}, and {Third} {Parties}.
\newblock {\em Criminal Justice Policy Review}, 9(3-4):279--306.

\bibitem[Riedel and Rinehart, 1996]{RiedelMurderClearancesMissing1996}
Riedel, M. and Rinehart, T.~A. (1996).
\newblock Murder {Clearances} and {Missing} {Data}.
\newblock {\em Journal of Crime and Justice}, 19(2):83--102.
\newblock Publisher: Routledge \_eprint:
  https://doi.org/10.1080/0735648X.1996.9721548.

\bibitem[Roberts, 2007]{RobertsPredictorsHomicideClearance2007}
Roberts, A. (2007).
\newblock Predictors of {Homicide} {Clearance} by {Arrest}: {An} {Event}
  {History} {Analysis} of {NIBRS} {Incidents}.
\newblock {\em Homicide Studies}, 11(2):82--93.
\newblock Publisher: SAGE Publications Inc.

\bibitem[Roberts and Lyons, 2011]{RobertsHispanicVictimsHomicide2011}
Roberts, A. and Lyons, C.~J. (2011).
\newblock Hispanic {Victims} and {Homicide} {Clearance} by {Arrest}.
\newblock {\em Homicide Studies}, 15(1):48--73.
\newblock Publisher: SAGE Publications Inc.

\bibitem[Roman, 2013]{RomanRaceJustifiableHomicide2013}
Roman, J.~K. (2013).
\newblock Race, {Justifiable} {Homicide}, and {Stand} {Your} {Ground} {Laws}:
  {Analysis} of {FBI} {Supplementary} {Homicide} {Report} {Data}.
\newblock Technical report, The Urban Institute, Washington, DC.

\bibitem[Rudin, 2019]{RudinStopexplainingblack2019a}
Rudin, C. (2019).
\newblock Stop explaining black box machine learning models for high stakes
  decisions and use interpretable models instead.
\newblock {\em Nature Machine Intelligence}, 1(5):206--215.

\bibitem[Sampson and Lauritsen, 1997]{SampsonRacialEthnicDisparities1997}
Sampson, R.~J. and Lauritsen, J.~L. (1997).
\newblock Racial and {Ethnic} {Disparities} in {Crime} and {Criminal} {Justice}
  in the {United} {States}.
\newblock {\em Crime and Justice}, 21:311--374.
\newblock Publisher: The University of Chicago Press.

\bibitem[Shapley, 1952]{ShapleyValueNPersonGames1952}
Shapley, L.~S. (1952).
\newblock A {Value} for {N}-{Person} {Games}.
\newblock Technical report, RAND Corporation.

\bibitem[Strauss, 2017]{Straussbetternabbingmurderers2017}
Strauss, B. (2017).
\newblock D.{C}. is better at nabbing murderers than a generation ago.
\newblock Technical report, DC Policy Center, Washington, DC.

\bibitem[Sturup et~al., 2015]{SturupUnsolvedhomicidesSweden2015}
Sturup, J., Karlberg, D., and Kristiansson, M. (2015).
\newblock Unsolved homicides in {Sweden}: {A} population-based study of 264
  homicides.
\newblock {\em Forensic Science International}, 257:106--113.

\bibitem[{The Washington Post}, 2018]{WashingtonPostMurderImpunity2018}
{The Washington Post} (2018).
\newblock Murder with {Impunity}.

\bibitem[{The Washington Post}, 2021]{TheWashingtonPostHowPostmapped2021}
{The Washington Post} (2021).
\newblock How {The} {Post} mapped unsolved murders.

\bibitem[Thebault et~al., 2021]{Thebault2020wasdeadliest2021}
Thebault, R., Fox, J., and Tran, A.~B. (2021).
\newblock 2020 was the deadliest gun violence year in decades. {So} far, 2021
  is worse.
\newblock {\em Washington Post}.

\bibitem[Wellford and Cronin, 1999]{WellfordAnalysisVariablesAffecting1999}
Wellford, C. and Cronin, J. (1999).
\newblock An {Analysis} of {Variables} {Affecting} the {Clearance} of
  {Homicides}: {A} {Multistate} {Study}.
\newblock Technical report, Justice Research and Statistics Association,
  Washington, DC.

\bibitem[Wellford et~al., 2019]{WellfordClearinghomicides2019}
Wellford, C.~F., Lum, C., Scott, T., Vovak, H., and Scherer, J.~A. (2019).
\newblock Clearing homicides.
\newblock {\em Criminology \& Public Policy}, 18(3):553--600.
\newblock \_eprint:
  https://onlinelibrary.wiley.com/doi/pdf/10.1111/1745-9133.12449.

\bibitem[Wolfgang, 1958]{WolfgangPatternsCriminalHomicide1958}
Wolfgang, M.~E. (1958).
\newblock {\em Patterns in {Criminal} {Homicide}}.
\newblock University of Pennsylvania Press, Philadelphia, PA.

\bibitem[Xu, 2008]{XuCharacteristicsHomicideEvents2008}
Xu, Y. (2008).
\newblock Characteristics of {Homicide} {Events} and the {Decline} in
  {Homicide} {Clearance}: {A} {Longitudinal} {Approach} to the {Dynamic}
  {Relationship}, {Chicago} 1966-1995.
\newblock {\em Criminal Justice Review}, 33(4):453--479.
\newblock Publisher: SAGE Publications Inc.

\bibitem[Štrumbelj and Kononenko,
  2014]{StrumbeljExplainingpredictionmodels2014}
Štrumbelj, E. and Kononenko, I. (2014).
\newblock Explaining prediction models and individual predictions with feature
  contributions.
\newblock {\em Knowledge and Information Systems}, 41(3):647--665.

\end{thebibliography}

\newpage

\appendix

\setcounter{equation}{0}
\setcounter{section}{0}
\setcounter{subsection}{0}
\setcounter{figure}{0}
\setcounter{table}{0}
\makeatletter
\renewcommand{\theequation}{S\arabic{equation}}
\renewcommand{\thefigure}{S\arabic{figure}}

\section{SUPPLEMENTARY MATERIALS}
\subsection*{Descriptive Statistics}

\begin{figure}[!hbt]
    \centering
    \includegraphics[scale=0.55]{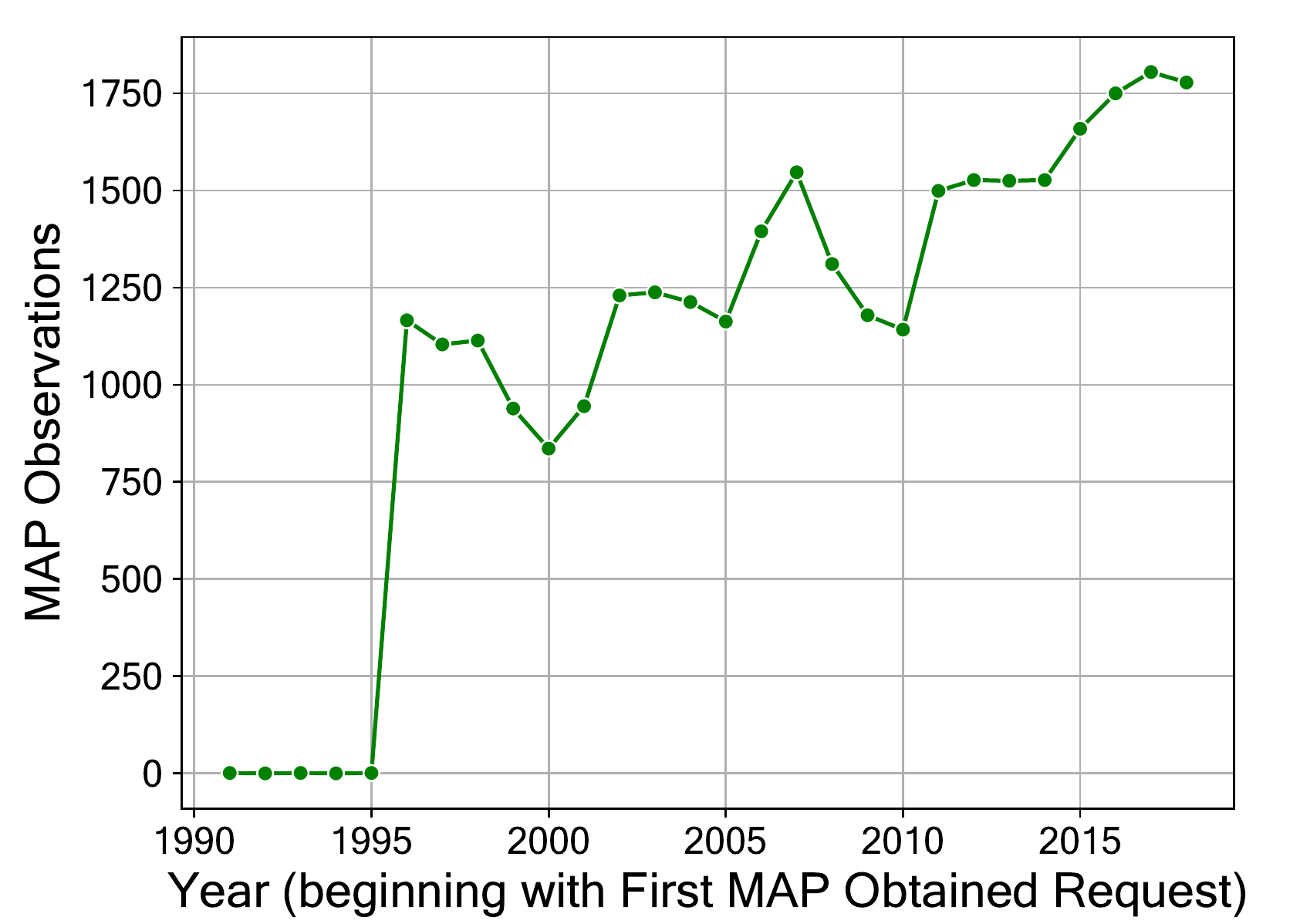}
    \caption{Yearly number of observations obtained by MAP through FOIA, starting from first year of first obtained record.}
    \label{fig:map}
\end{figure}

\begin{figure}[!hbt]
    \centering
    \includegraphics[scale=0.5]{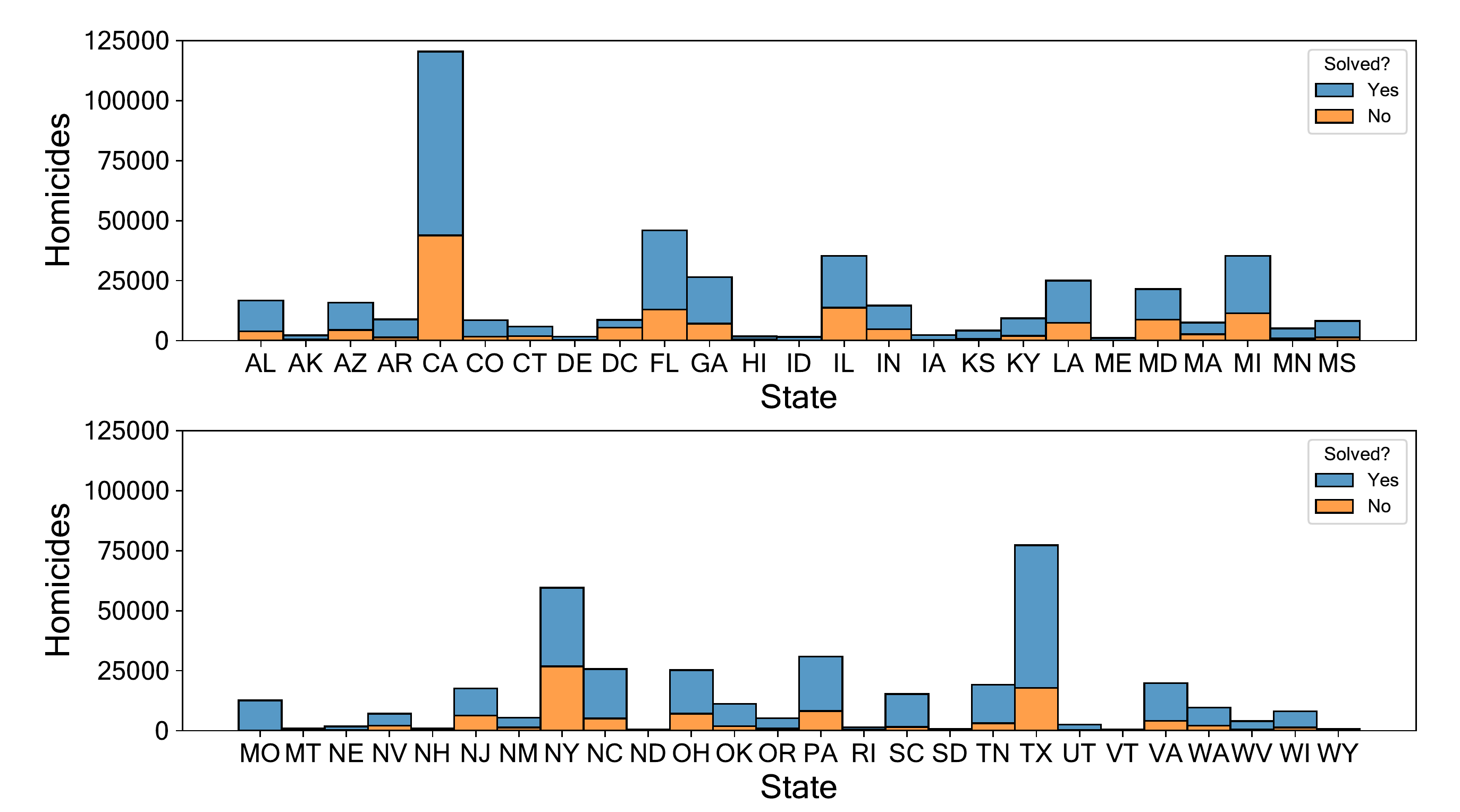}
    \caption{Bar chart for the number of homicides for each state in the MAP dataset. Colors indicate the proportion of solved and unsolved cases.}
    \label{fig:desc_cases}
\end{figure}

Figure \ref{fig:map} shows the yearly number of observations in the MAP dataset that have been obtained directly by MAP, and therefore were not initially recorded in the SHR files.
Figure \ref{fig:desc_cases} highlights the distribution of homicides represented in the MAP dataset, divided by state, with colors mapping the number of solved and unsolved events. California (120,462) is by far the state with the highest number of recorded homicides, 1.55 times more than the number of the second-ranked state, Texas (77,226). The third is the state of New York, with 59,516 homicides. The states with the lowest number of cases are North Dakota (457), Vermont (518), South Dakota (620), and Wyoming (824). The average number of homicides per state is 15,779.43, with a standard deviation of 21,612.23, indicating that the distribution is right-skewed.

In terms of the percentage of solved cases, the average is 77.01\%, with a standard deviation of 11.07. The states with the highest percentage of solved cases are North Dakota (94.31\%), South Dakota (93.06\%), Montana (91.54\%), and Idaho (90.73\%). Notably, these are among the least populated states in the US. The states with the lowest percentage instead are the District of Columbia, with a negative record of 37.91\%, followed by New York (55.02\%), Maryland (59.15\%), Illinois (61.13\%) and California (63.63\%). 

\subsection*{Algorithmic Approaches}\label{algoapp}
Nine different algorithmic approaches have been compared to choose the best-performing model at the national level. Furthermore, for each algorithm, relevant hyperparameters have been grid-searched, increasing the number of candidates to optimize the predictive ability of each of the tested approaches. The Scikit Learn API has been used for all the models described below \citep{BuitinckAPIdesignmachine2013}.

\paragraph{Ridge Regression}
Ridge Regression (also called L2 Regression) represents an alternative to the OLS model designed to perform regularization on the coefficient estimates to reduce the variance (and therefore instability) of estimated coefficients \citep{HastieElementsStatisticalLearning2009}. It is particularly useful when covariate coefficients are highly correlated. While very similar to OLS, Ridge Regression solves a model having the loss function equal to the linear least squares function, with regularization provided by the L2-norm, governed by a tuning parameter $\lambda$ that quantifies the extent of the penalty for each coefficient. Ridge shrinks estimators towards 0, with higher penalties leading to coefficients closer to 0. A total of 18 Ridge regression models have been fit, with three different types of solvers (i.e., newton-cg, liblinear, saga) and six different $C$ values, which map the inverse of the regularization parameter $\lambda$ , meaning that smaller values implying stronger regularization (i.e., 0.01, 0.5, 1, 5, 10, 50). 
\paragraph{Lasso Regression}
Lasso logistic regression is another popular regularization method that, differently from Ridge, uses the L1-norm, thus allowing coefficients to be equal exactly to 0, therefore involving a variable selection method as coefficients equal to 0 are implicitly removed from the model \citep{HastieElementsStatisticalLearning2009}. For this reason, Lasso is generally preferred when higher model interpretability is the goal. The $\lambda$ parameter also controls the penalization strength. A total of twelve Lasso regression models were compared, using two solvers (i.e., liblinear and saga) and six $C$ values (i.e., 0.01, 0.5, 1, 5, 10, 50).

\paragraph{Elastic Net Regression}
Elastic Net (EN) is a third regularization method for regression models that uses both the penalties seen in Ridge and Lasso regression \citep{HastieElementsStatisticalLearning2009}. Besides the $\lambda$ parameter, which governs the strength of regularization, Elastic Net also has a $\alpha$ parameter which varies from 0 to 1, with 0 transforming the penalty function in the Ridge term, and 1 reducing it to the Lasso term. A value between 0 and 1 is therefore required to optimize Elastic Net. Six total models have been compared for Elastic Net regression, using Saga as the solver, 0.5 as the $\alpha$ parameter and six $C$ values to map regularization strength (i.e., 0.01, 0.5, 1, 5, 10, 50).
\paragraph{Decision Trees}
Decision Trees (DT) is a very popular non-parametric method used in supervised learning \citep{QuinlanInductiondecisiontrees1986}. It is engineered through a recursive partition to predict the value of a target variable via rules learned from the feature space. Decision Trees are represented as trees with directed edges, where all the (internal) nodes are split into multiple subspaces based on feature values. Among its advantages are its high interpretability and low computational cost. However, Decision Trees can suffer from overfitting and low ability in generalization, as well as sensitivity to outliers. 
\paragraph{Random Forests}
Random Forests (RF) is an ensemble method that seeks to solve the disadvantages displayed by Decision Trees \citep{BreimanRandomForests2001c}. Random Forests address such problems by growing many classification trees, with each tree giving a specific prediction such that the forest then uses the most common prediction as the final one. Due to their ensemble nature, Random Forests often perform very well in complex tasks, especially compared to simpler alternative methods. Beyond higher accuracy, among the pros of Random Forests are its efficiency, the unbiased estimation of coefficients, and the elimination of the overfitting problem. Among the important hypeparameters associated with Random Forests are the criterion used to evaluate the quality of each node split, the number of estimators (namely, the number of grown trees), and the trees' depth included in the forest. For the present application, a total of 48 Random Forests models have been compared, using both the Gini and Entropy criteria for node splitting, maximum tree depth equal to 5,10,20,30,50, and 100 and four different number of estimators, i.e., 100, 200, 500, and 700.
\paragraph{Gradient Boosting Machines}
Gradient Boosting Machine (GBM) is also an ensemble method that has gained momentum in the last decades due to the high performance demonstrated in various contexts \citep{FriedmanGreedyfunctionapproximation2001}. Its application encompasses both regression and classification problems. Different from RF, which consists of ensembles of independent trees, Gradient Boosting Machines generates an ensemble of weak learners in the form of trees, in which each successive tree improves the performance of the previous. Among the most important hyperparameters (upon which the literature suggests to find a trade-off), are the number of successive estimators and the learning rate, which maps the shrinkage level of each tree, or more in general the size of the steps made by the algorithm to optimize the loss function, with smaller learning rate leading to higher computational cost, but more control in finding minima. A total of 18 GBMs have been compared, using three number of trees (50, 10, 200) and six learning rates (0.001, 0.01, 0.1, 0.3, 0.5).

\paragraph{Linear Support Vector Machines}
Support Vector Machines (SVM) is a supervised learning algorithm that can be employed for solving both regression and classification tasks \citep{Bosertrainingalgorithmoptimal1992}. It is a non-probabilistic algorithm that seeks to find the hyperplane dividing a given set of data into two classes, and it possesses some relevant advantages that have contributed to its popularity as well as success in machine learning. Particularly, SVM is effective with high dimensional spaces and is memory efficient given that it employs a subset of training points, called support vectors, to improve its performance. SVM also includes regularization in both L1 and L2 forms. A Linear SVM has been employed in the present study using a liblinear solver. A total of 12 candidates have been compared, resulting from the combination of two different penalties (L1 and L2) and six C values (a hyperparameter mapping the strength of the chosen regularization, as seen in LASSO and Ridge Regression), namely 0.01, 0.5, 1, 5, 10, 50.

\paragraph{XGBoost}
XGBoost (XGB), which stands for ``Extreme Gradient Boosting'', is an algorithm designed to handle supervised learning tasks, which is substantially an expansion (or, arguably, an improvement) of Gradient Boosting Machines \citep{ChenXGBoostScalableTree2016}. Compared to GBM, however, XGBoost has important distinguishing features. First, it is engineered to optimize memory usage and minimize computational cost, enhancing, for instance, parallelization. Second, XGBoost uses regularization, thus often leading to better generalization. Third, XGBoost handles missing value internally and automatically, reducing data processing costs on the researcher's side. Among the most important hyperparameters that have to be evaluated when using XGBoost, are the number of estimators and the learning rate, similarly to the procedure in GBM. Additionally, XGB also has a Gamma parameter for regularization. For modeling purposes, a total of 36 XGBoost candidates have been tested as a combination of three different number of estimators (50, 100, 200), four different learning rates (0.01, 0.1, 0.3, 0.5), and three gamma regularization terms (0, 0.5, 1). 

\paragraph{Linear Discriminant Analysis}
Linear Discriminant Analysis (LDA) is a linear classification technique that is used for classification and dimensionality reduction purposes and works for both binary and multi-class tasks \citep{HastieElementsStatisticalLearning2009}. Significantly simpler than other models, LDA assumes that features are normally distributed, and that variance is equal across each feature. Through these two assumptions, LDA projects data into a hyperplane to maximize separation between classes finding a boundary, specifically maximizing the distance between means of the two and minimizing the variation existing between them. Two hyperparameters have been tuned to test its predictive ability, namely the type of solver and the shrinkage parameter. Concerning the former, the three available solvers in scikit-learn have been used, namely singular value decomposition, least-square, and eigenvalue decomposition. Concerning the latter, two levels of shrinkage (no shrinkage and automatic shrinkage through Ledoit-Wolf lemma) led to a total of six LDA candidates.

\subsection*{Performance Evaluation }\label{perfo}
Different metrics exist to evaluate the predictive performance of a machine learning classifier, yet some metrics are specifically well-suited for particular conditions or problems. Given that, as anticipated, the distribution of solved and unsolved homicides is not equal in national and local terms, the two metrics that have been considered are Balanced Accuracy and Precision.

\paragraph{Balanced Accuracy}
Balanced Accuracy is calculated as the arithmetic mean of sensitivity (i.e., the ratio between true positives and the sum of true positives and false negatives) and specificity (i.e., the ratio between true negatives and the sum of true negatives and false positives). The value ranges from 0 to 1, with 0.5 indicating that the classifier has a predictive ability that is not above chance. The formula is:

\begin{equation}
\begin{aligned}
    Balanced \: Accuracy = \frac{Sensitivity+Specificity}{2}=\\\frac{(\frac{True \:Positives}{True\:Negatives+False\:Negatives})+(\frac{True\:Negatives}{True\:Negatives+False\:Positives})}{2}
\end{aligned}
\end{equation}

\paragraph{Precision}
Mathematically, Precision is calculated as the ratio between true positives and the sum of true positives and false positives and can take any value from 0 to 1, with higher values indicating higher Precision. Mathematically:
\begin{equation}
    Precision = \frac{True \: Positives}{(True \: Positives+False\: Positives)}
\end{equation}

\subsection*{Grid Search Model Results}
\begin{figure}[!hbt]
\centering
\begin{subfigure}[b]{0.70\textwidth}
   \includegraphics[width=1\linewidth]{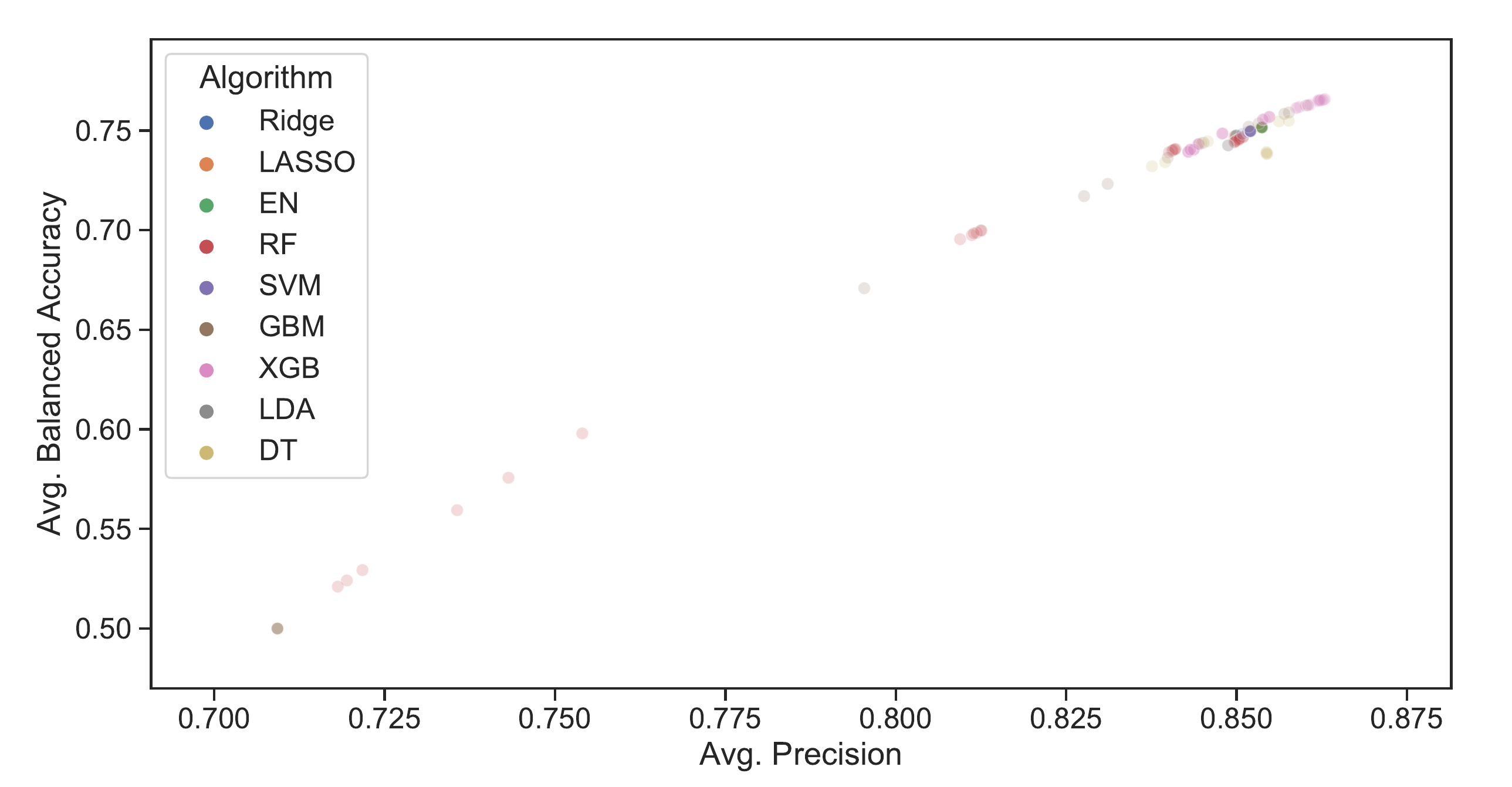}
   \caption{}
   \label{fig:NgA} 
\end{subfigure}

\begin{subfigure}[b]{0.70\textwidth}
   \includegraphics[width=1\linewidth]{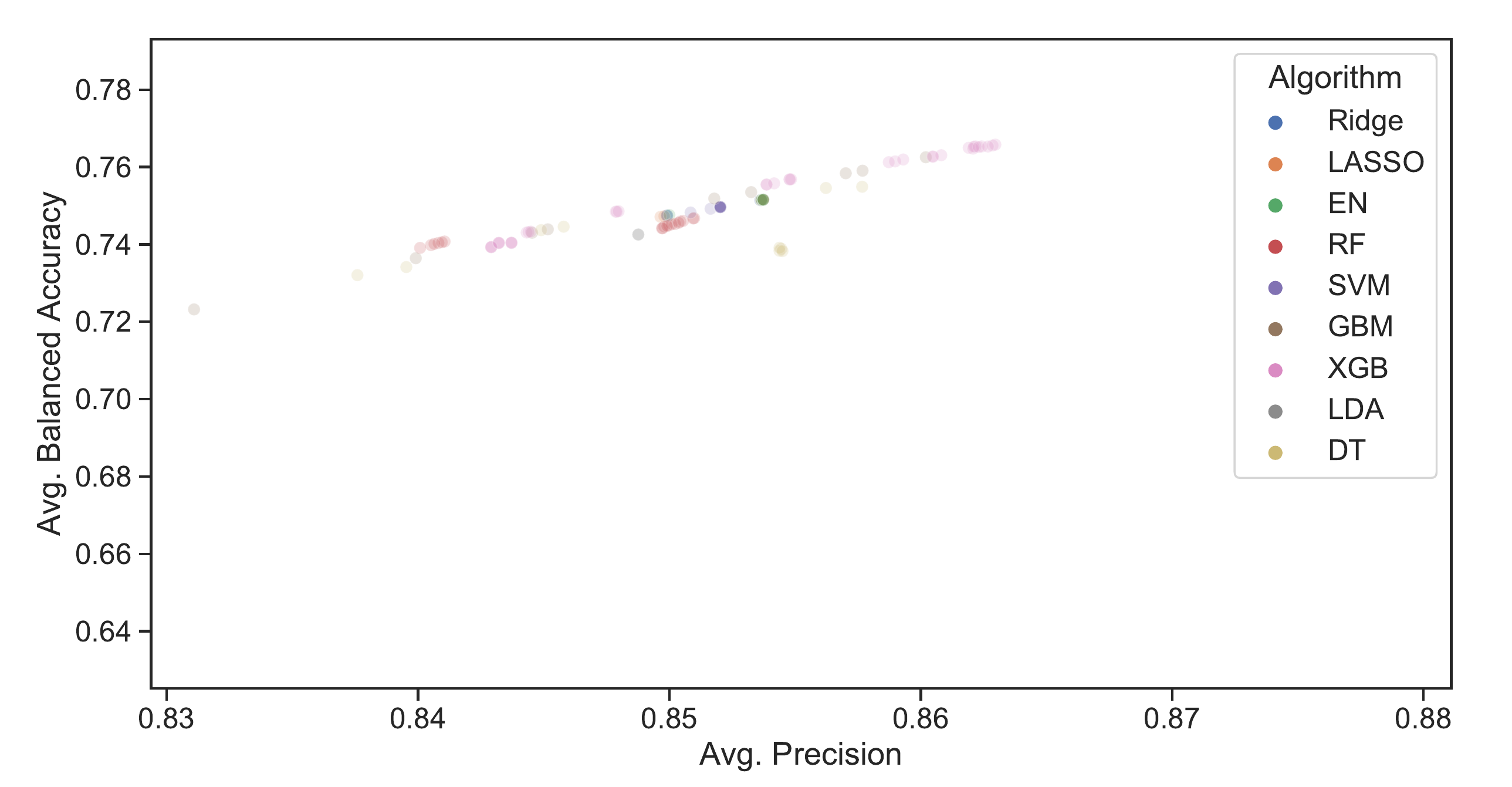}
   \caption{}
   \label{fig:NgB}
\end{subfigure}
\caption{Scatterplot showing the performance of all tested algorithmic models and configurations at the national level. (a) displays the entire distribution, while (b) offers a zoomed visualization of the models and configurations with the highest performance overall.}
\end{figure}

Figure \ref{fig:NgA} displays a scatterplot to assess the relationship between average Balanced Accuracy and average Precision for all the cross-validated models tested through grid-search of hyperparameters for optimization. Figure \ref{fig:NgB} shows a zoomed snapshot of the scatterplot in \ref{fig:NgA}, to appreciate the distribution of the best performing models. Both plots demonstrate that, although with slight margins compared to some alternatives, XGBoost obtains the highest performance in both metrics.

\begin{figure}[!hbt]
    \centering
    \includegraphics[scale=0.45]{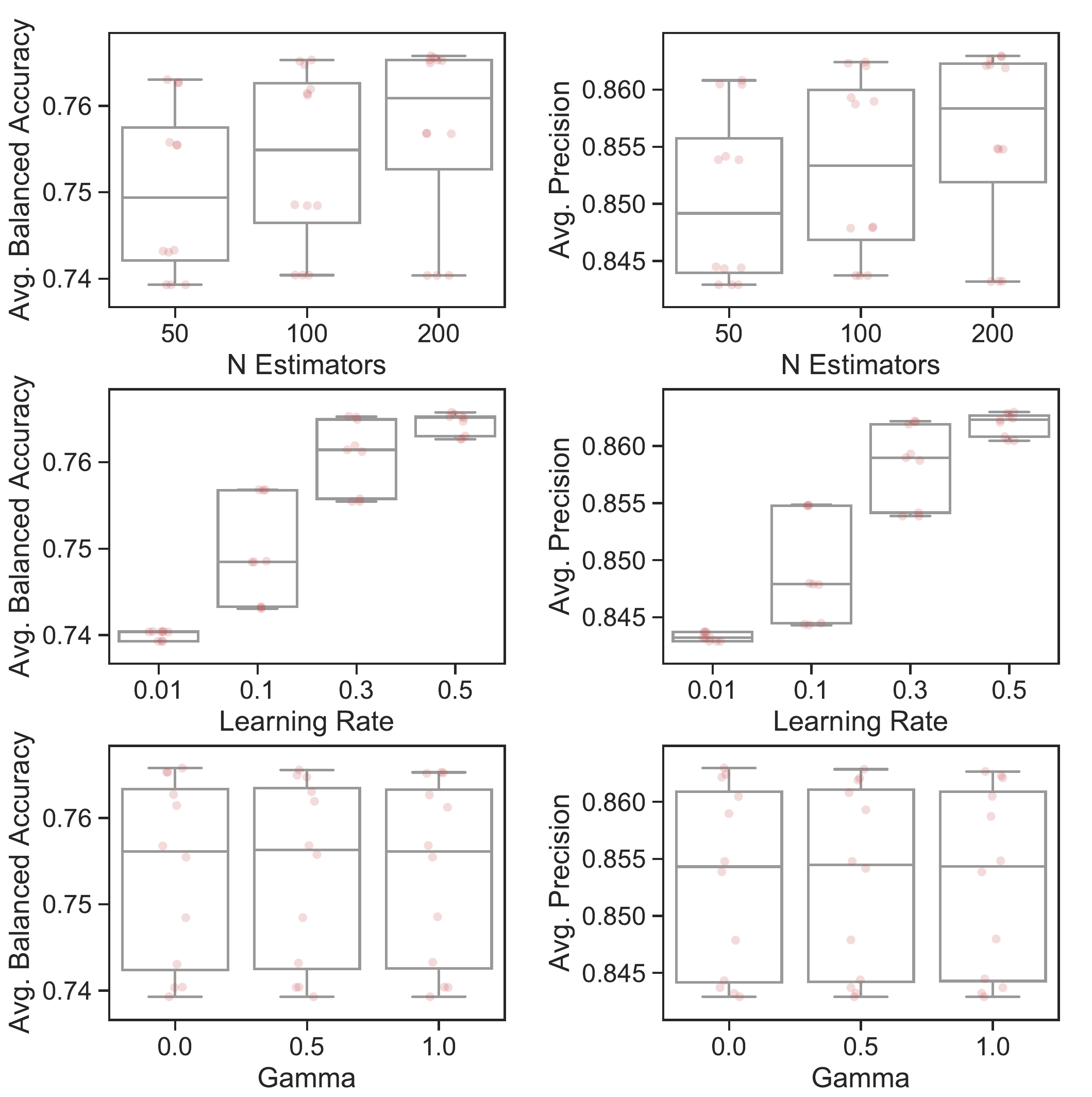}
    \caption{Distribution of predictive performance of XGBoost configurations, with focus on each different hyperparameter.}
    \label{fig:xgbnatpar}
\end{figure}

\newpage
Figure \ref{fig:xgbnatpar} offers a visualization of performance in terms of Balanced Accuracy and Precision across cross-validated models using data in the training set to show variations based on each hyperparameter used in the grid search. The plot refers to XGBoost, the algorithm that achieved the highest performance among all. First, it should be noted that a higher number of estimators guaranteed higher predictive accuracy. 

Somehow contrarily to most recommendations to keeping it low, setting a learning rate equal to 0.5 had slightly better results than the second-best, i.e., 0.3, and more sensible improvements compared to the other three alternatives. 

In terms of regularization, no differences were found between regularized and non-regularized models. In light of this, the final model used no regularization.

\subsection*{Model Explainability}

\subsubsection*{National level: Local Explanation}
\begin{figure}[!hbt]
\centering
\parbox{7.5cm}{
\includegraphics[width=7.5cm]{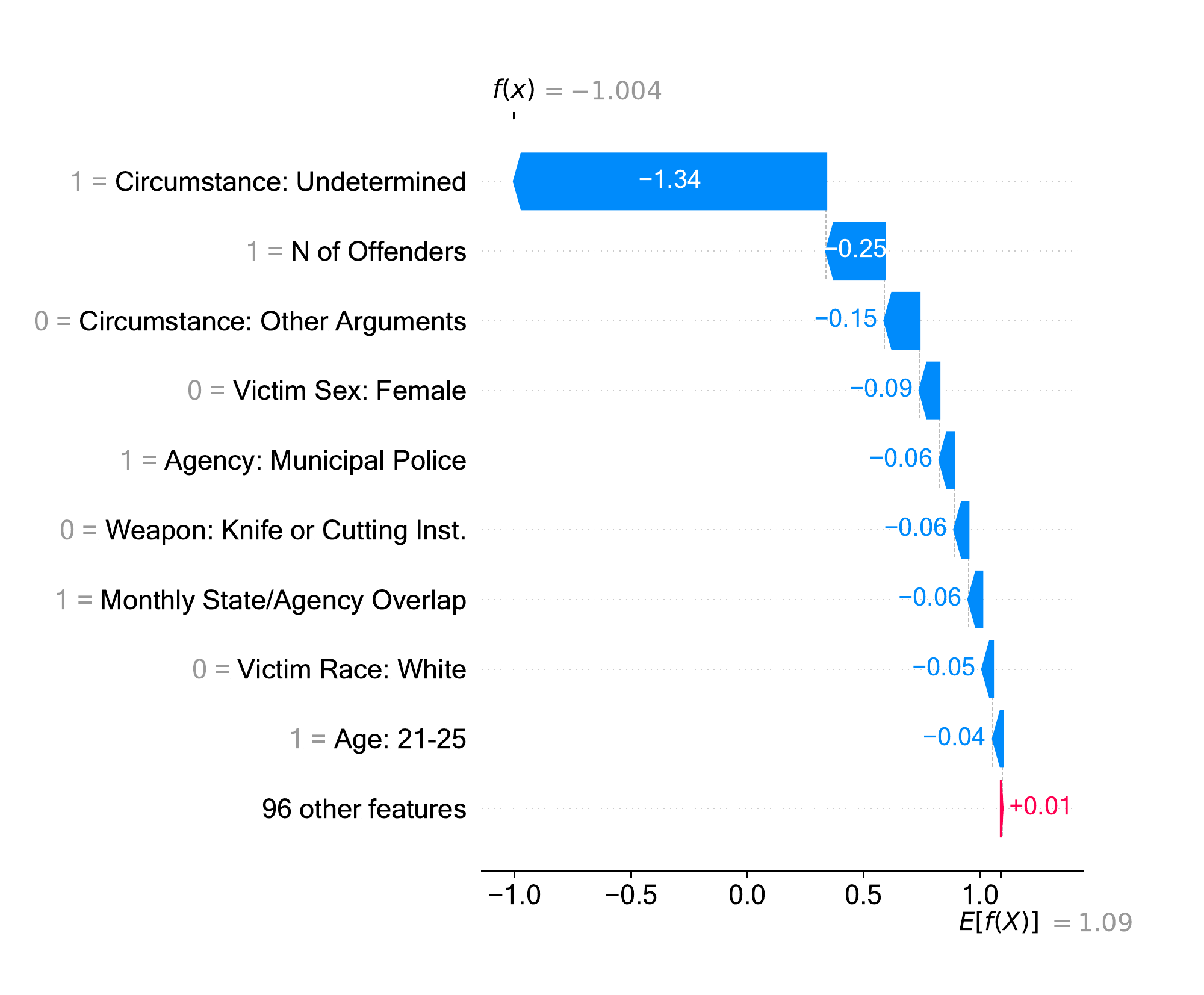}
\caption{Local Prediction Explanation for 1st Observation in the Test Set (Coefficients are Reported in Log Odds)}
\label{fig:2figsA}}
\qquad
\begin{minipage}{7.5cm}
\includegraphics[width=7.5cm]{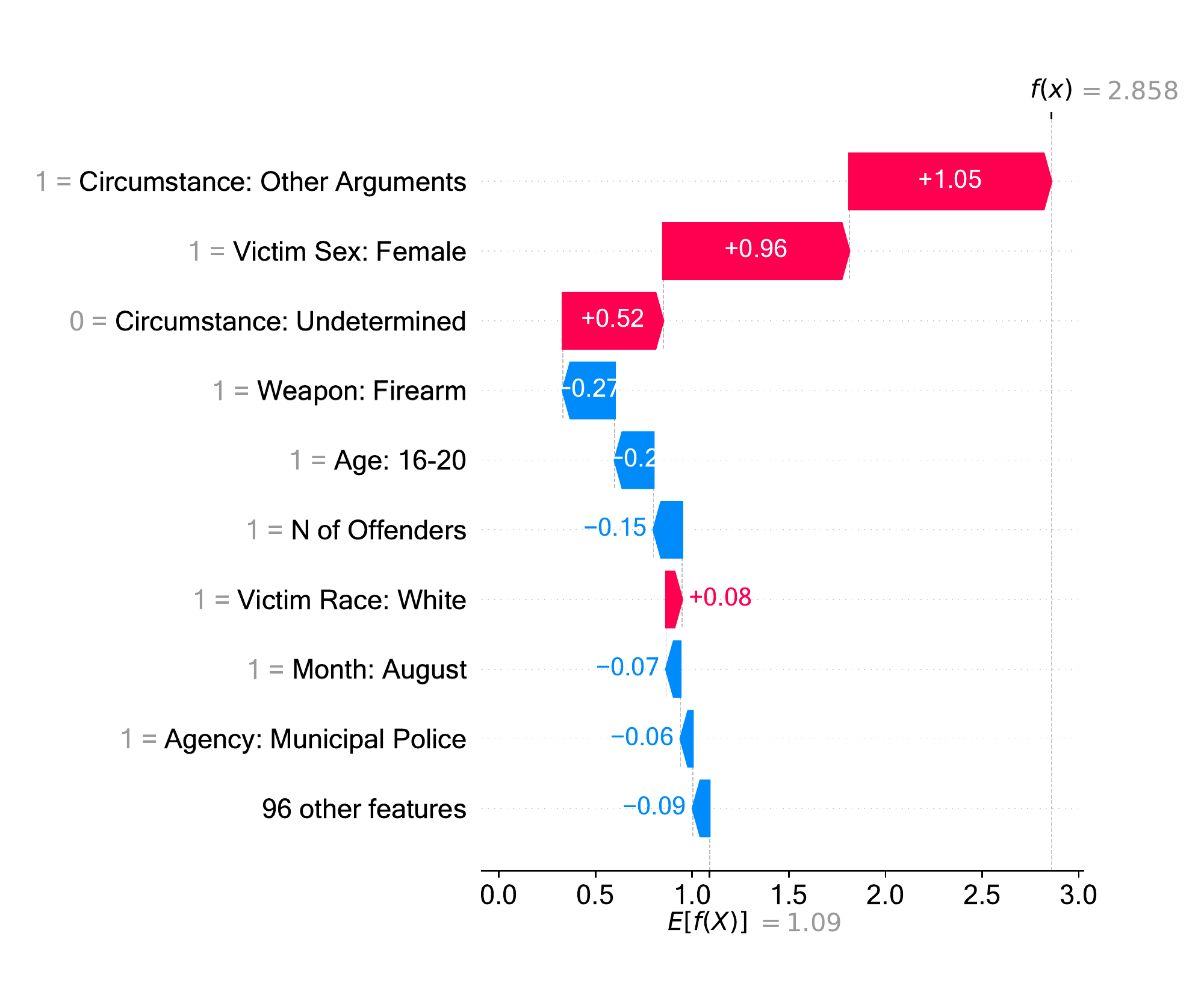}
\caption{Local Prediction Explanation for 100th Observation in the Test Set (Coefficients are Reported in Log Odds)}
\label{fig:2figsB}
\end{minipage}
\end{figure}

Figures \ref{fig:2figsA} and \ref{fig:2figsB} provides two examples of local explanations for single predictions. In particular, the plots visualize each feature's outcomes and impact for the 1st and 100th observations in the national test set. The baseline prediction is 1.09 Log Odds, meaning that there is a probability that, without considering any feature, an homicide is solved with probability of 0.748, because, if we transform the log odds in odds by taking its exponential, it becomes $Odds=exp(Log\:Odds=2.974$, and therefore $p=Odds/(1+Odds)=0.748$.
In the first example, the final prediction is equal to -1.004, amounting to a probability of $0.26$ that the homicide is solved, with the contribution of the first nine most important features being all negative. For instance, given that the homicide is linked to undetermined circumstances, the log odds that the case is solved are reduced by 1.34 compared to the baseline. Similarly, given that the victim is not a female, the probability decreases by additional 0.09 log odds. 

In the second example, being the baseline probability equal to the previous (trivially), the final prediction equals 2.858 log-odds, which indicates a probability of 0.945 that the homicide was solved. This prediction is achieved by considering the impact of each feature. For instance, the fact that the homicide was perpetrated with a firearm leads to a reduction of 0.22 log odds in the probability that the homicide is solved, but at the same time, given that the circumstances refer to "other arguments", the log-odds increase by 1.05. Also, the victim is a female, and this raises the likelihood of a solved homicide. Overall, the combination of features associated with this observation reaches a very high probability of having a cleared homicide.

\subsubsection*{State-level focus on Race as a Relevant Feature}

\begin{figure}[!hbt]
    \centering
    \includegraphics[scale=0.4]{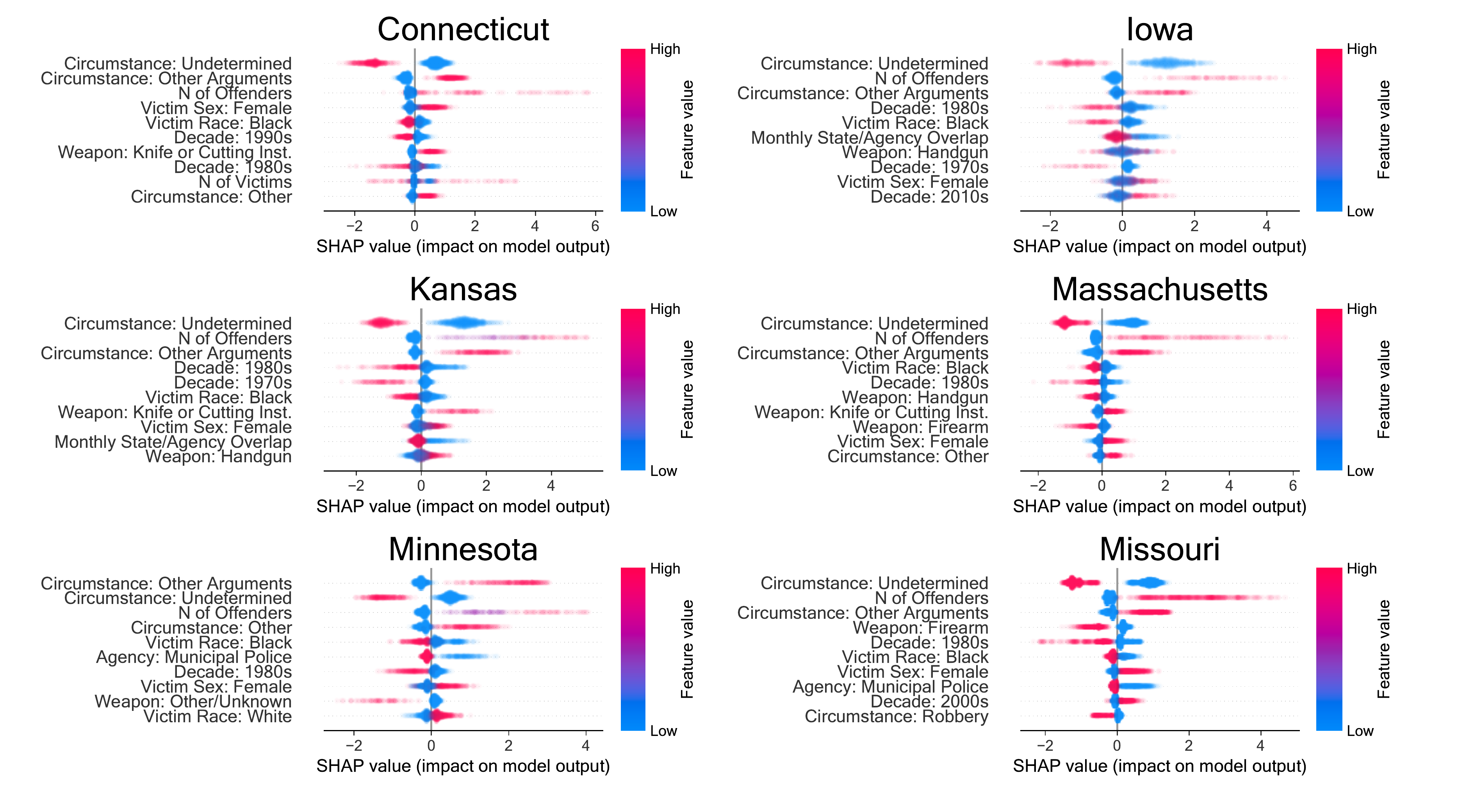}
    \caption{Distribution of SHAP values for the top 10 features at the state level for Connecticut, Iowa, Kansas, Massachusetts, Minnesota, and Missouri, a sample of the states in which homicides with Black victims are less likely to be solved.}
    \label{fig:black_sup}
\end{figure}

Figure \ref{fig:black_sup} expands the focus on a sample of states in which homicides with black victims have a lower probability of being solved. These include Connecticut (N=5,864), Iowa (N=2,373), Kansas (4,185), Massachusetts (7,538), Minnesota (5,045), and Missouri (19,723). Michigan also had the same pattern in the six states with the highest number of homicides. 

In these six states, the findings are clear and, in the case of Minnesota, are also coupled with the concomitant positive role of "Victim Race: White" on predictions, reinforcing disparities among racial groups.

\subsection*{Matching with Washington Post Data}\label{wp}
\subsubsection*{Process}
To evaluate the accuracy of the "Solved?" variable, I have employed data from the Unsolved Homicide Database of the "Murder with Impunity" project of \cite{WashingtonPostMurderImpunity2018} (WP dataset henceforth). The project gathered data on 52,179 homicides that occurred from 2007 to 2017 in 50 of the largest US cities and are available at \href{https://github.com/washingtonpost/data-homicides}{https://github.com/washingtonpost/data-homicides} \citep{TheWashingtonPostHowPostmapped2021}. These data were retrieved by the Washington Post through a variety of sources, and provided several information on each case, along with the outcome of the investigation. 

To match the MAP and the WP datasets, I have used information on the following overlapping variables:

\begin{itemize}
    \item City (labeled as "Agency" in the MAP dataset, and "city" in the WP dataset)
    \item Year (labeled as "Year" in the MAP dataset, and derived from "reported\_date" in the WP dataset)
    \item Month (labeled as "Month" in the MAP dataset, and derived from "reported\_date" in the WP dataset)
    \item Victim's Age (labeled as "VicAge" in the MAP dataset, and "victim\_age" in the WP dataset)
    \item Victim's Sex (labeled as "VicSex" in the MAP dataset, and "victim\_sex" in the WP dataset)
    \item Solved (labeled as "Solved" in the MAP dataset, and "disposition" in the WP dataset).
\end{itemize}
In each of the two datasets and for each observation I have created a code based on the selected five variables, which is processed as: 

\begin{equation}
    \text{Code(i)=Year(i)-Month(i)-City(i)-Victim's Age(i)-Victim's Sex(i)}
\end{equation}

This process led to matching a total of 27,450 homicides that were both in the MAP and WP datasets. The differences in the match (52.6\% of total homicides present in the WP dataset) are likely due to imprecise reporting of the victim's age (supporting the decision to use a categorical variable for transforming it as done in the main models presented in the article) or location of occurrence.

After comparing the "Solved" outcomes in the two, 21,910 homicides had equal outcomes (79.81\% of the total). A total of 2,782 were labeled as solved in the WP dataset while being unsolved in the MAP dataset (10.13\% of the total). This estimate is in line with those provided by \cite{HargroveMurderAccountabilityProject2019}. The project's website, in fact, signals that the dataset is around 90-95\% accurate in the determination of investigations' outcomes.

\subsubsection*{Robustness Checks: Predictive Explanations}

For robustness purposes, I have used the dataset (labeled as WP-MAP) containing the 27,450 homicides that I matched from both datasets in two different ways. First, I used the 21,910 homicides that had equal outcomes in both datasets to fit a SHAP model through XGBoost (with 200 learners and a learning rate equal to 0.5) to evaluate stability in relevant features.

Second, I have used all the 27,450 homicides and modified the outcome in the MAP dataset according to the one found in the WP one to ensure that the stability of important features holds even when partially modifying the distribution of the target variable. This means that if a homicide was labeled as "Unsolved" in the WP dataset, but was "Solved" in the MAP one (and vice versa), the value was modified to reflect the information in the WP dataset.

 These two versions have been constructed through the same feature processing steps carried out for the general MAP dataset, except for the exclusion of the "Decade" variable, given that all homicides in the combined WP-MAP dataset occurred between 2007 and 2017.

\begin{figure}[!hbt]
    \centering
    \includegraphics[scale=0.3]{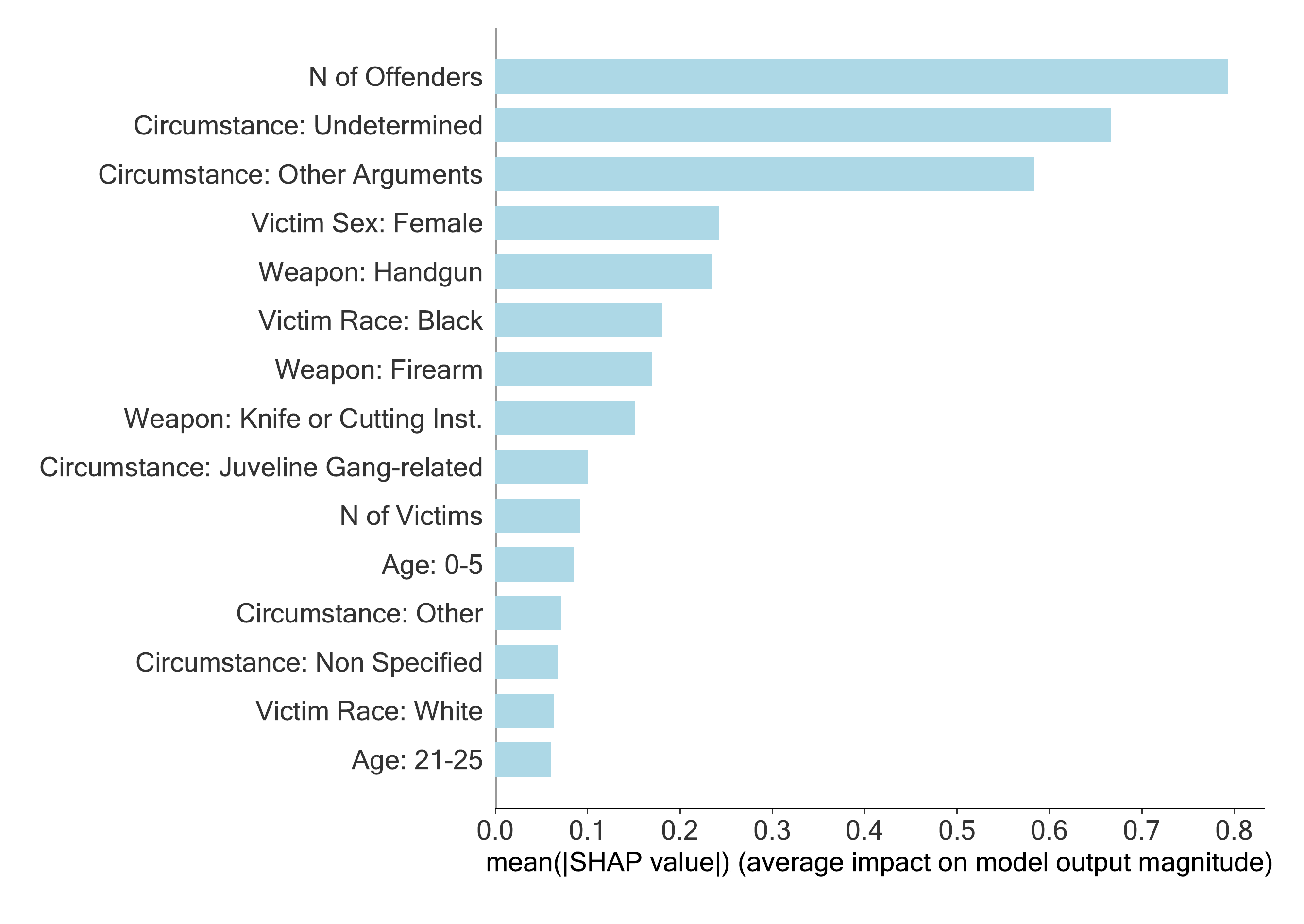}
    \caption{Robustness check: list of top 15 features with highest average SHAP value in absolute terms, mapping average impact on WP-MAP model output magnitude (only observations with equal outcomes, i.e. Solved or not, in both datasets were kept.}
    \label{fig:robu_shap_abs}
\end{figure}

\begin{figure}[!hbt]
    \centering
    \includegraphics[scale=0.35]{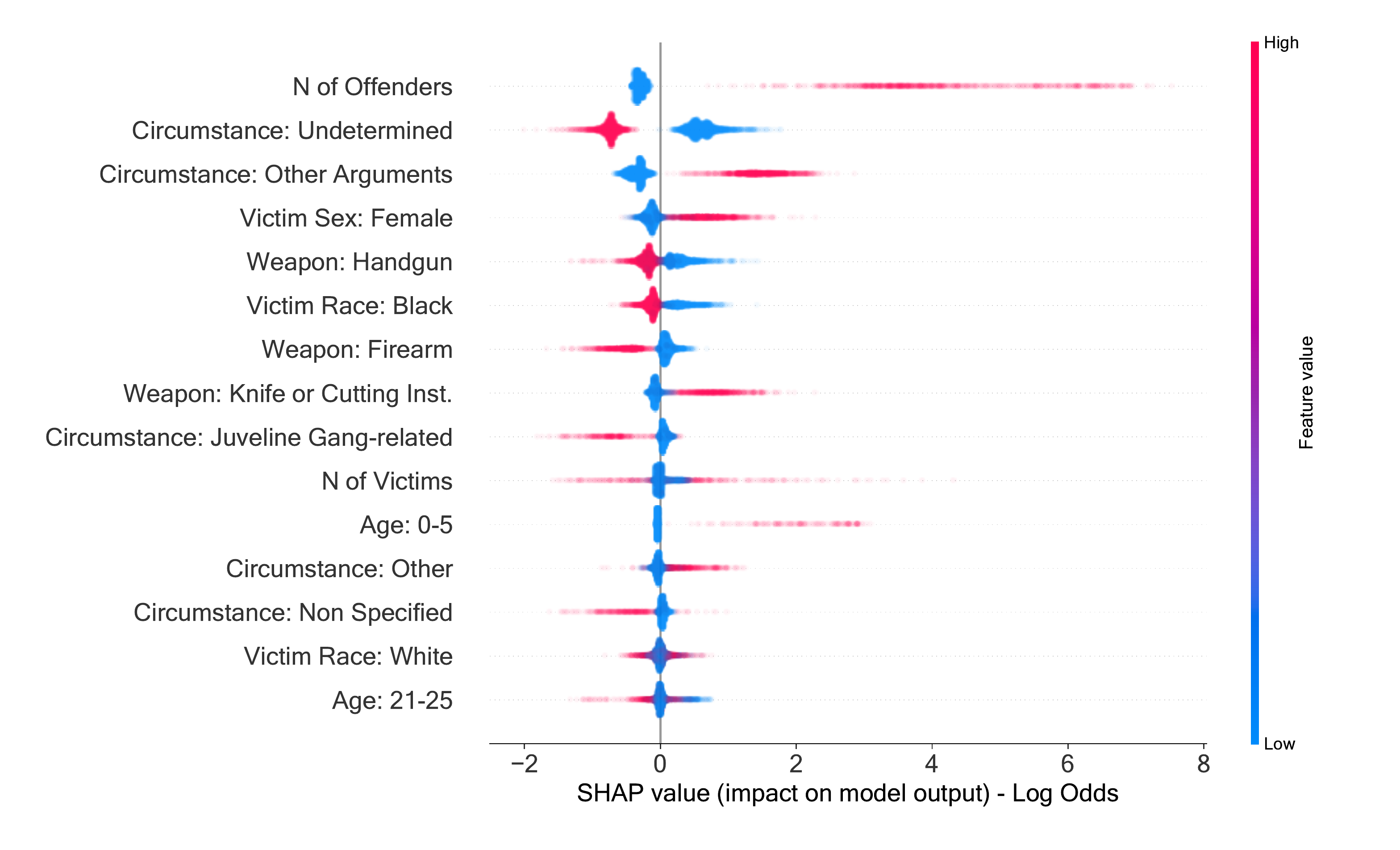}
    \caption{Robustness check: distribution of SHAP values for the top 15 most impactful features for the WP-MAP dataset (only observations that in both datasets had equal outcomes, i.e., Solved or not, were kept). Each point represents an observation in the US test set. Points falling left of the zero threshold mean a negative relationship with cleared homicides, while points falling right indicate otherwise. Each point is colored based on the value of the feature: in the case of a binary variable, if the instance is equal to 1 for a particular observation, the point will be colored in red, blue otherwise. Nuanced colors will identify count variables, e.g., Number of Offenders. }
    \label{fig:robu_shap_dist}
\end{figure}

\begin{figure}[!hbt]
    \centering
    \includegraphics[scale=0.3]{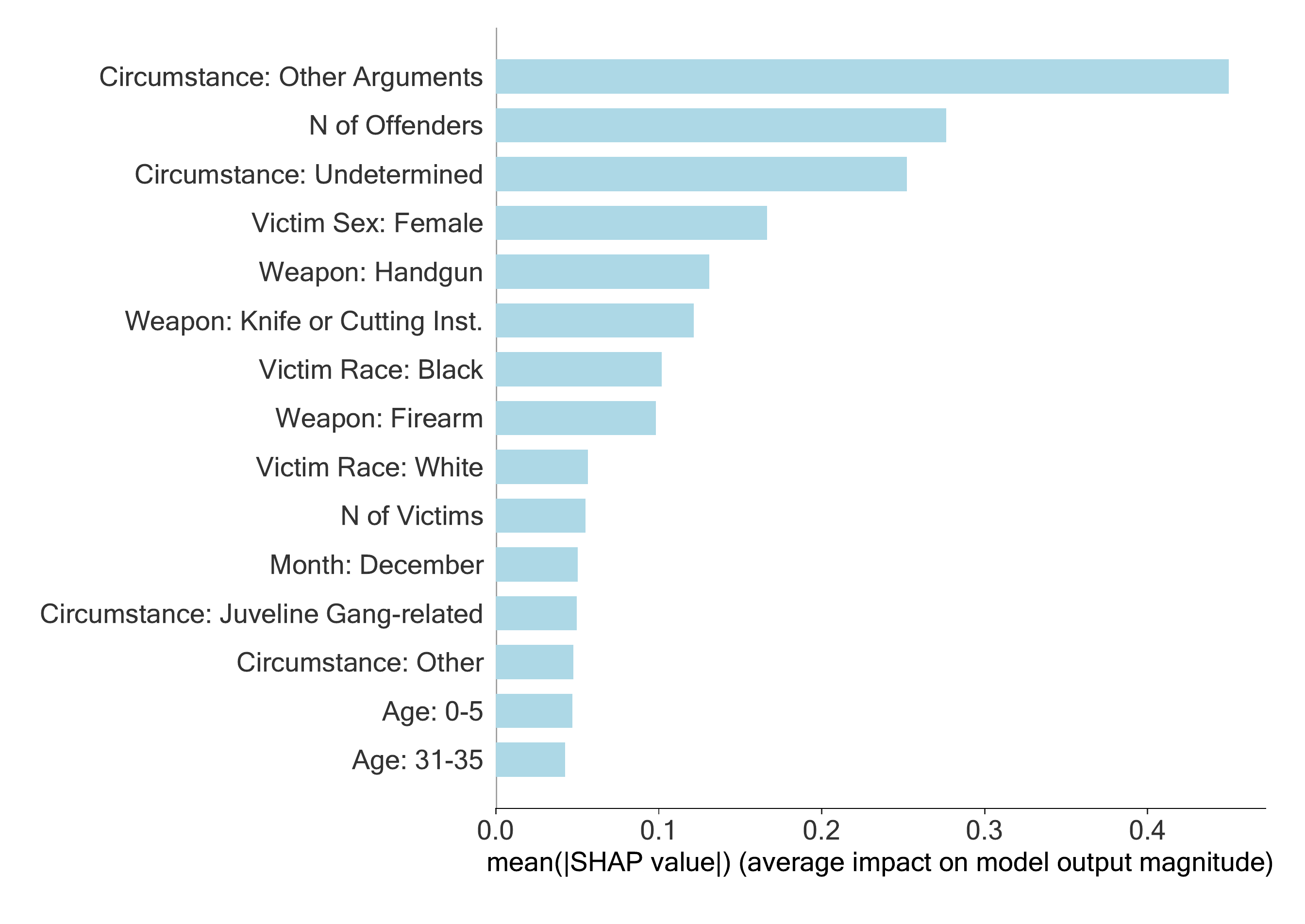}
    \caption{Robustness check: list of top 15 features with highest average SHAP value in absolute terms, mapping average impact on WP-MAP model output magnitude (target outcomes modified according to WP dataset).}
    \label{fig:robu2_shap_abs}
\end{figure}

\begin{figure}[!hbt]
    \centering
    \includegraphics[scale=0.35]{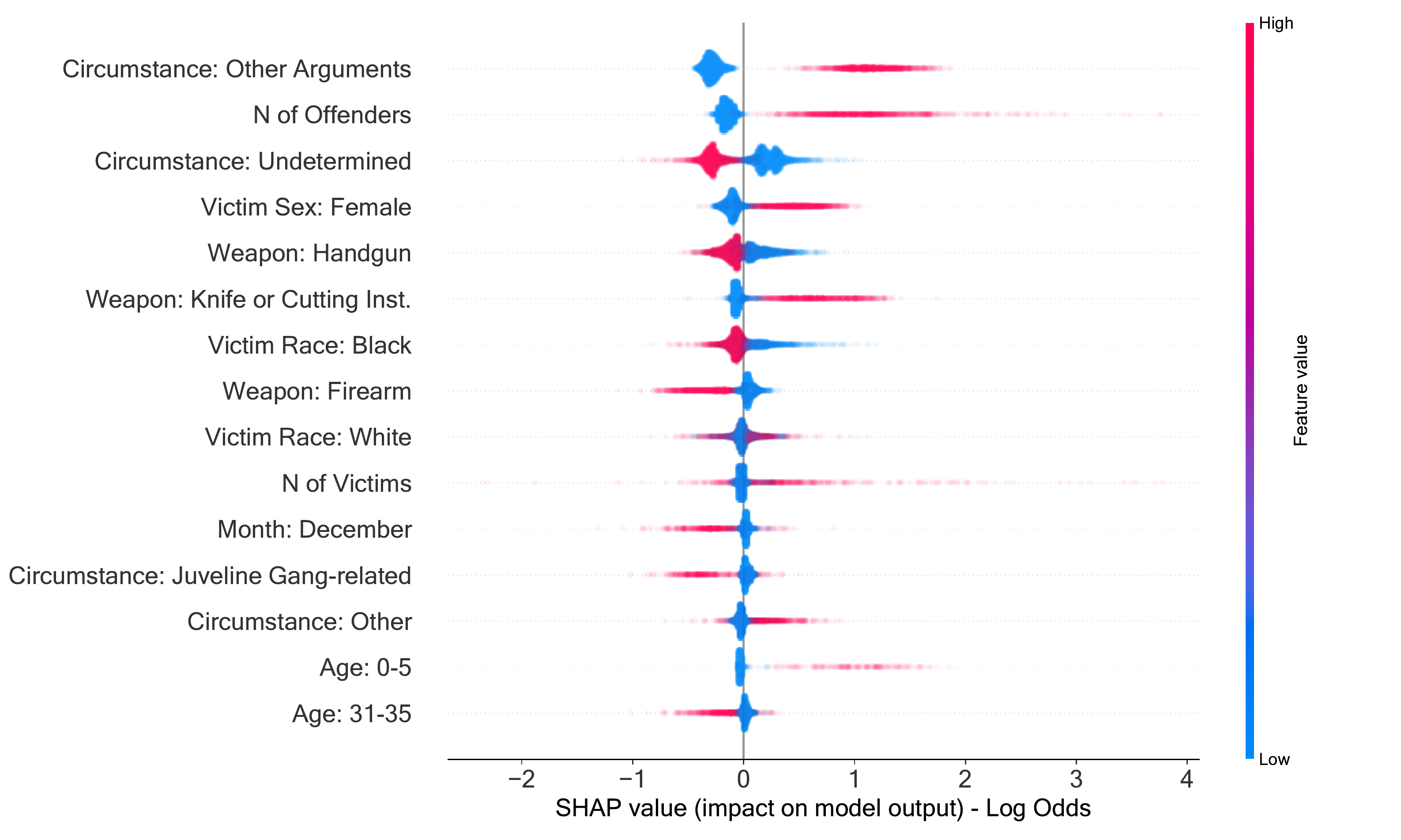}
    \caption{Robustness check: distribution of SHAP values for the top 15 most impactful features for the WP-MAP dataset (target outcomes modified according to WP dataset). Each point represents an observation in the US test set. Points falling left of the zero threshold mean a negative relationship with cleared homicides, while points falling right indicate otherwise. Each point is colored based on the value of the feature: in the case of a binary variable, if the instance is equal to 1 for a particular observation, the point will be colored in red, blue otherwise. Nuanced colors will identify count variables such as Number of Offenders. }
    \label{fig:robu2_shap_dist}
\end{figure}

Below are reported the SHAP results for the 15 features with the highest average impact on model output magnitude in both cases (Figures \ref{fig:robu_shap_abs} and \ref{fig:robu2_shap_abs}), as well as the distribution of the raw SHAP values for the same 15 features (Figures \ref{fig:robu_shap_dist} and \ref{fig:robu2_shap_dist}).

The outcomes (in terms of overall relevant features, as mapped by average |SHAP| values, and in terms of SHAP value distributions) align with findings obtained in the national-level analyses presented in the main text. Except for some variations (that can be partially explained by the absence of features mapping "Decades"), most of the relevant features overlap across different settings, even when modifying the MAP dataset outcomes according to the Washington Post dataset. Among the most relevant differences, is the greater importance that "Victim Race: Black" has on model outcomes. While it was already signaled as an important predictor at the national and state levels, homicides with Black victims are even more strongly associated with uncleared cases when considering the restricted WP-MAP datasets.

\clearpage

\newpage

\end{document}